\pdfoutput=1

\documentclass[11pt]{article}

\usepackage[]{acl}

\usepackage{times}
\usepackage{latexsym}

\usepackage[T1]{fontenc}

\usepackage[utf8]{inputenc}

\usepackage{microtype}

\usepackage[skip=0pt]{caption}
\usepackage{subcaption}
\usepackage{graphicx}
\usepackage{amsmath}
\usepackage{amssymb}
\usepackage{pifont}
\usepackage[dvipsnames]{xcolor}
\newcommand{\cmark}{\ding{51}}
\newcommand{\xmark}{\ding{55}}
\usepackage{booktabs}
\usepackage{algorithm}
\usepackage{algpseudocode}
\algrenewcommand\algorithmicrequire{\textbf{Input:}}
\algrenewcommand\algorithmicensure{\textbf{Output:}}
\usepackage{algpseudocode}
\usepackage{threeparttable}
\usepackage{multirow}
\usepackage{tabularx}
\usepackage{cleveref}
\usepackage{tcolorbox}
\usepackage{soul}
\usepackage{multirow}
\usepackage{colortbl}
\usepackage{transparent}
\usepackage{xspace}
\usepackage{paralist}
\usepackage{enumitem}

\colorlet{lightgray}{White!30!lightgray}
\colorlet{lightblue}{White!70!MidnightBlue}

\newcommand{\decorateSettingName}[1]{\textcolor{ForestGreen}{\textsf #1}\xspace}

\newcommand{\settingA}{\decorateSettingName{FewShot}}
\newcommand{\settingB}{\decorateSettingName{FewShotU}}
\newcommand{\settingC}{\decorateSettingName{FewShotUP}}

\newcommand{\direct}{\textbf{Direct}\xspace}
\newcommand{\pmi}{\textbf{PMI}\xspace}
\newcommand{\pdo}{\textbf{PDO}\xspace}

\newcommand {\mm}[1] {\ifmmode{#1}\else{\mbox{\(#1\)}}\fi}
\newcommand{\Dcal}        {\mm{\mathcal{D}}}
\newcommand{\Xcal}        {\mm{\mathcal{X}}}
\newcommand{\Ycal}        {\mm{\mathcal{Y}}}

\crefname{figure}{Fig.}{Figs.}
\crefname{section}{Sec.}{Secs.}
\crefname{equation}{Eqn.}{Eqns.}
\crefname{appendix}{Appx.}{Appx.}
\crefname{table}{Table}{Tables}

\usepackage{graphicx}

\graphicspath{{./figures/}}

\title{In-Context Example Ordering Guided by Label Distributions}

\author{Zhichao Xu\textsuperscript{1,2} \quad
Daniel Cohen\textsuperscript{3} \quad
Bei Wang\textsuperscript{1,2} \quad
Vivek Srikumar\textsuperscript{1} \\
\\
\textsuperscript{1}Kahlert School of Computing, University of Utah \\
\textsuperscript{2}Scientific Computing and Imaging Institute, University of Utah \\
\textsuperscript{3}Dataminr, Inc. \\
{\tt  zhichao.xu@utah.edu}\\
}

\begin{document}
\maketitle
\begin{abstract}
By allowing models to predict without task-specific training, in-context learning (ICL) with pretrained LLMs has enormous potential in NLP. However, a number of problems persist in ICL. In particular, its performance is sensitive to the choice and order of in-context examples. Given the same set of in-context examples with different orderings, model performance may vary between near random to near state-of-the-art.
In this work, we formulate in-context example ordering as an optimization problem. We examine three problem settings that differ in the assumptions they make about what is known about the task. Inspired by the idea of learning from label proportions, we propose two principles for in-context example ordering guided by model's probability predictions. We apply our proposed principles to thirteen text classification datasets and nine different autoregressive LLMs with 700M to 13B parameters. We demonstrate that our approach outperforms the baselines by improving the classification accuracy, reducing model miscalibration,  and also by selecting better in-context examples.

\end{abstract}

\section{Introduction}
\label{sec:intro}

An intriguing property of large language models like the GPT~\cite{brown2020language,openai2023gpt} and PaLM families of models~\cite{chowdhery2022palm,anil2023palm} 
is their ability to ``learn in context''. That is, the model can achieve competitive predictive performance with only a task description and a few training examples with no parameter updates~\cite{brown2020language,min-etal-2022-rethinking,xie2021explanation}.
Model predictions can sometimes even match full fine-tuning performance~\cite{lu-etal-2022-fantastically}. 

In-context learning (ICL)---the idea of prompting LLMs with only a few examples, also known as few-shot prompting---has shown promise across NLP.
Yet, many problems persist with this paradigm. 
Prior work has shown that ICL is sensitive to different natural language instructions and different orderings of in-context  examples~\cite{sorensen-etal-2022-information,lu-etal-2022-fantastically}. 
Merely changing the ordering of a fixed set of examples can change the predictive performance from that of nearly fully-tuned models to random guessing. 
\citet{lu-etal-2022-fantastically} studied in-context ordering and proposed heuristics to select the performant orderings. 
However, prior work on example ordering assumes (to different degrees) that an additional dataset is available to help reorder the in-context examples. 

We ask: \emph{Can we select the best in-context example orderings  \textbf{with no labeled} data beyond the in-context ones?}
We draw inspiration from the idea of learning from label distributions~\cite{yu2014learning,dulac2019deep}, which shows that the prior probability distributions of labels can weakly supervise label predictors. 
We build upon this insight to improve the quality of in-context predictions, and in particular, to select performant in-context example orderings.

\begin{figure}[!t]
\begin{subfigure}{0.9\columnwidth}
    \includegraphics[width=\linewidth]{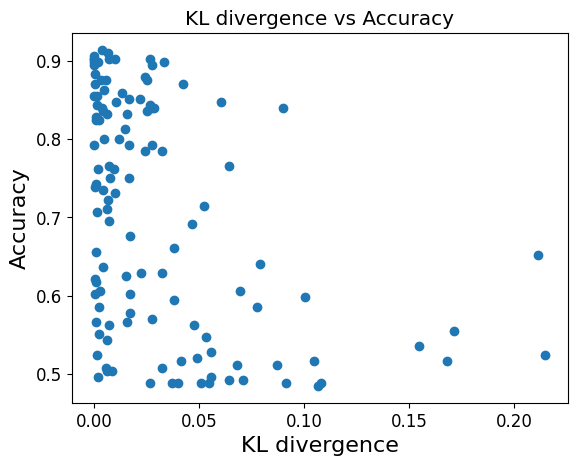}
    \caption{\settingA}
    \label{subfig:sst2-settingA}
\end{subfigure}
\begin{subfigure}{0.9\columnwidth}
    \includegraphics[width=\linewidth]{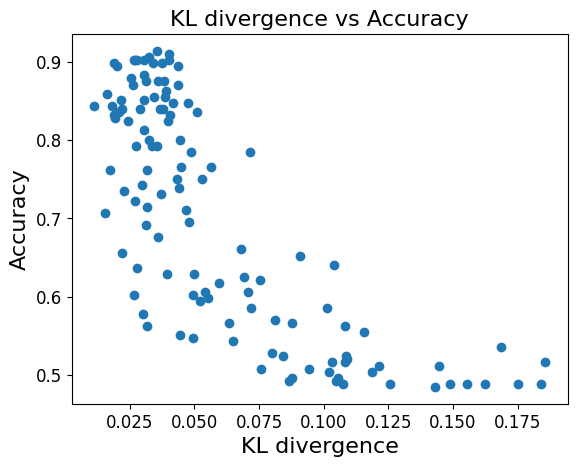}
    \caption{\settingC}
    \label{subfig:sst2-settingC}
\end{subfigure}
\caption{
KL-divergence vs accuracy for \settingA and \settingC on SST-2 dataset, with a backbone language model OPT-1.3B. 
}
\label{fig:teaser}
\end{figure}

We consider two cases: (a) when we only have in-context examples (\settingA), and (b) when we also have unlabeled examples (\settingB) and additionally know the prior label distributions (\settingC). 
In all cases, we only use the model's output probability distributions over candidate outputs.  
These distributions serve as a direct indicator of the model's confidence as well as the bias carried from pretraining and in-context examples. 

Given a set of in-context examples, we propose to select the best ordering that, on the corpus level, has a probability distribution over candidate labels, such that it is 
\begin{inparaenum}[(a)]
\item less biased towards certain labels, or,
\item close to a prior label distribution, if known. 
\end{inparaenum}

\Cref{fig:teaser} illustrates the two criteria using OPT-1.3B as a backbone language model.
Each point corresponds to a certain ordering of a fixed set of in-context examples. 
\cref{subfig:sst2-settingA} corresponds to case (a). 
Its x-axis is the KL-divergence between the uniform distribution and the model's probability for the null input given in-context examples. 
This KL-divergence captures the model's bias towards certain labels; smaller values indicate less bias. 
\cref{subfig:sst2-settingC} corresponds to case (b). 
Its x-axis is KL-divergence between model's average probability distribution over unlabeled samples and the prior label distribution.
In \cref{subfig:sst2-settingA}, accuracy is weakly inversely correlated with KL-divergence, indicating performant orderings tend to be less biased towards certain labels. 
In \cref{subfig:sst2-settingC}, the negative correlation is stronger, indicating the marginal label probabilities of performant orderings tends to be close to informative priors.

Our approach, \emph{Probability Distribution Ordering} (\pdo), effectively improves in-context predictions on 13 text classification datasets and 9 language models with 700M--13B parameters. 
It not only improves the classification accuracy and reduces variance across all datasets and models, but also improves models' confidence calibration, making them more suitable for real-world deployment. 

Finally, in analysis experiments, we study how well \pdo can select in-context examples for a task.
Prior work on task-level in-context example selection
requires \emph{labeled} development data~\cite{chang-jia-2023-data,nguyen2023context}.
We  show that \pdo improves task-level example selection, matching CondAcc~\cite{chang-jia-2023-data} without the need for a labeled development set.

\section{Background \& Notation}
\label{sec:background}

We seek to order in-context examples to improve both predictive accuracy and model calibration. 
This section reviews in-context learning (ICL) and model calibration, and introduce relevant notation. Through the paper, we use the word ordering and permutation interchangeably.  

\subsection{In-Context Learning}
\label{sec:ICL}

Consider the task of predicting a label $y \in \Ycal$ for an input $x \in
\mathcal{X}$, where $\mathcal{X}$ and $\Ycal$ denote the textual input space 
and the label space, respectively. 
The label $y$ can be \emph{verbalized} into a natural language token. 
For example, for a sentiment classification task, the input space may be product reviews, and the label space $\Ycal=\{ +, -\}$ may be verbalized to the words \texttt{positive} and \texttt{negative}. 

In-context learning naturally applies to the few-shot setting. We have a small set of $k$ training examples $x_i$ paired with corresponding labels $y_i$, denoted by $\Dcal=\{ (x_1, y_1), (x_2, y_2), \ldots, (x_k,y_k) \}$. 
To predict the label for a new example $x$, we construct an input for a language model by concatenating a certain ordering $\pi(\Dcal)$ of these $k$ examples with $x$. 
Using this input $(\pi(\Dcal),x)$, the model generates a probability distribution $ P(y\mid \pi(\Dcal), x)$ over the label set $\Ycal$ via the verbalized variants of each label $y \in \Ycal$. 
For a classification task, the predicted label for $x$ is therefore $\arg \max_{y \in \Ycal} P(y\mid \pi(\Dcal),x)$.
Since the label for $x$ is predicted using a probability distribution directly obtained from the pretrained language model without further processing, we follow prior works~\citep[e.g.,][]{min-etal-2022-noisy} and refer to this approach as the {\direct} method. 

\citet{holtzman-etal-2021-surface} proposed an alternative scoring function to \direct, where they scored each label $y \in \Ycal$ for the unseen example $x$ as: 
\begin{equation}
    \text{PMI}(x,y) = \log \frac{P(y \mid \pi(\Dcal), x)}{P(y \mid \pi(\Dcal), \text{null}).}
    \label{eq:pmi}
  \end{equation}
Here, the score $\text{PMI}(x,y)$ denotes the pointwise mutual information between a label $y$ and input $x$. 
In practice, $P(y \mid \pi(\Dcal),\text{null})$ requires simply setting input $x$ to an empty string.
The predicted label is now $\arg \max_{y \in Y} \text{PMI}(x, y)$.
The intuition is that a higher PMI value indicates a stronger association between the input $x$ and a candidate label $y$. We refer to this approach   as the {\pmi} method.

The above scoring functions are agnostic to in-context example order. 
However, recent works~\cite[e.g.,][]{lu-etal-2022-fantastically,wu-etal-2023-self} have shown that ICL performance is sensitive to the orderings of in-context examples. 
To address this issue, \citet{lu-etal-2022-fantastically} assumed a development set and presented a heuristic for ordering the prompts. 
Their heuristic (and also the approaches we present) can be used with both the \direct and \pmi methods.

\subsection{Confidence Calibration}
\label{sec:confidence-calibration}

Previous work on ICL have mainly evaluated their results with performance metrics such as accuracy for classification tasks. 
However, models can grow over-confident about their predictions, which is problematic for deployment.
Prior works in calibrating neural networks~\citep[e.g.][]{guo2017calibration} have argued that a network should provide a calibrated \emph{confidence measure} with its prediction.
Specifically, the mean probability of a correct prediction for $x$ should be equal to its average accuracy, e.g., all predictions at 70\% confidence level should have an average accuracy of 70\%. 
A model's confidence calibration can be measured by the expected difference between its confidence and accuracy, 
\begin{equation}
    \underset{\hat{p}}{\mathbb{E}} \big[ |P(\hat{y}=y|\hat{p}=p)-p | \big].
    \label{eq:confidence_calibration}
\end{equation}
Here,  $\hat{y}$ denotes the predicted label, $\hat{p}$ denotes the associated confidence.
In practice, this is often measured by Expected Calibration Error~\cite[ECE,][]{naeini2015obtaining}. 
ECE approximates~\cref{eq:confidence_calibration} by partitioning predictions into a number of equally-spaced bins and taking a weighted average of the bins’ accuracy-confidence difference.

Despite being effective in terms of performance, \pmi skews the output
probability distribution, leading to miscalibrated model outputs. 
In~\cref{eq:pmi}, if the denominator $P(y_i\mid\pi(\Dcal),\text{null})$ is already skewed by context $\pi(\Dcal)$, it can magnify the skewness of output probability distribution.

Consider the following example: for a binary sentiment classification task and a new input   example $x$, the probability distribution over \{\texttt{positive}, \texttt{negative}\} is $(0.7,0.3)$ for input $(\pi(\Dcal), x)$, and $(0.3,0.7)$ for input $(\pi(\Dcal), \text{null})$, respectively. 
Taking softmax over the PMI-adjusted scores according to~\cref{eq:pmi} yields a final probability distribution of $(0.92, 0.08)$; whereas the model still predicts $x$ as \texttt{positive}, it may have grown over-confident.
In~\cref{sec:experiments}, we show empirically that \pmi leads to higher miscalibration than \direct.

\section{A Proposal for Ordering Selection}
\label{sec:method}

We seek to find an ordering of the $k$ in-context examples $\mathcal{D}$ that has the best predictive performance and leads to calibrated probabilities.
To do so, we need to rank the $k!$ permutations and select the highest performing one. However, since we are operating in the few-shot setting, it is important to specify the task information  we are allowed to use.

Prior efforts~\cite{lu-etal-2022-fantastically,wu-etal-2023-self,sorensen-etal-2022-information} on in-context example ordering operate under different resource settings, making it difficult to perform comprehensive comparisons.
We study three settings, in the order of increased information:
\begin{enumerate}
\item  \settingA: Only a few (typically $8$ to $32$) labeled in-context examples $\Dcal$ are available.
\item  \settingB:  In addition to a few labeled in-context examples $\Dcal$, an \emph{unlabeled} development set $X$ is available. 
\item  \settingC: In addition to in-context examples $\Dcal$  and an unlabeled development set $X$, we know the prior \emph{probability distribution} $Q(\Ycal)$ over the label space $\Ycal$. 
\end{enumerate} 
In the \settingC setting, the prior  over labels may be determined from prior information (e.g.,~previous experiments) or a subjective expert assessment (e.g.,~the probability of a certain disease in the population assessed by a clinician). 

The \settingB setting---where we do not know the prior label distribution---can be seen as a special case of \settingC that uses an \emph{uninformative (or flat)  prior}, i.e.,~a uniform prior distribution of $Q(\Ycal)=\text{Unif}(\Ycal)$ over the development set.
We therefore consider two cases separately: 
\begin{inparaenum}[(1)]
\item when we only have the in-context examples (\settingA); and
\item when we also have an unlabeled set of examples (\settingB) and additionally know the prior label distribution (\settingC).
\end{inparaenum}

\subsection{\settingA with Only In-Context Examples}
\label{sec:fewshot}

In this setting, we have no  information about the task beyond the in-context examples. Thus, any label predictor should be maximally uncertain when presented with no inputs. 
That is, a good in-context example ordering should lead to the model being unbiased towards certain labels with a null input (e.g., an empty string). 
We state our first principle: 
\begin{tcolorbox}[width=\linewidth]
\vspace{-5pt}
\textsc{Principle I}: When unlabeled examples are not available, well-ordered in-context examples should lead to the probability distribution of a null input having the minimum KL divergence to a uniform distribution.
\vspace{-5pt}
\end{tcolorbox}
\noindent
\textsc{Principle I} can be instantiated as a function that scores an ordering $\pi:=\pi(\Dcal)$ as follows:
\begin{equation}
    \mathcal{L}(\pi) = D_{\text{KL}} \left( P(\Ycal \mid \pi,\text{null}) \,||\, \text{Unif}(\Ycal) \right).
    \label{eq:kl_div_null}
\end{equation}

\subsection{\settingB and \settingC}

We have unlabeled examples in \settingB, and also the prior label distribution in \settingC. 

Consider the prior distribution $Q$ over the label space $\Ycal$. 
The distribution $Q$ can be obtained by marginalizing out the input space  $\Xcal$ as: 
\begin{align}
Q(y) & = \sum_{x \in \Xcal} P_{\Ycal \mid \Xcal}(y \mid x) P_\Xcal(x) \label{eqn:sum}\\
     & = E_\Xcal \left[ P_{\Ycal \mid \Xcal}(y \mid x)\right]. \label{eqn:expectation}
\end{align} 
The probability $P_\Xcal$ in \cref{eqn:sum} denotes the \emph{unknown} distribution over the input space $\Xcal$. 
The probability $P_{\Ycal \mid \Xcal}$ in \cref{eqn:sum} is the label distribution conditioned on the input $\Xcal$. 
It is the object of study and is provided to us by the language model. 

Therefore, if we have access to the unlabeled set $X$ sampled i.i.d. from the natural data distribution, we can approximate the expectation in  \cref{eqn:expectation} as an empirical mean $\hat{Q}(y) \approx Q(y)$: 
\begin{align}
\label{eqn:empirical-mean}
\hat{Q}(y) = \frac{1}{|X|}\sum_{x \in X} P(y \mid x).
\end{align}

Now, suppose we know the prior distribution $Q$ and have access to the \emph{unlabeled} set $X$. 
Using $X$ and any ordering $\pi:=\pi(\Dcal)$ of the in-context examples, we can compute $\hat{Q}$ of \cref{eqn:empirical-mean} and measure its difference from $Q$. Concretely, we define the \emph{observed label distribution} $\hat{P}$ in terms of the model-induced label  distributions: 
\begin{align}
\label{eqn:empirical-mean-2}
\hat{P}(y\mid \pi) = \frac{1}{|X|}\sum_{x\in X}P(y\mid \pi, x). 
\end{align}

Now, we can state our second principle: 
\begin{tcolorbox}[width=\linewidth]
\vspace{-5pt}
\textsc{Principle II}: Given an unlabeled set of examples and the prior label distribution, well-ordered in-context examples should produce an observed label distribution that matches the prior label distribution.
\vspace{-5pt}
\end{tcolorbox}
\noindent
\noindent\textsc{Principle II} gives us a function that scores  a permutation $\pi$ as follows:
\begin{equation}
    \mathcal{L}(\pi) = D_{\text{KL}} \left( \hat{P}\left(\Ycal |\pi\right) \,||\, Q(\Ycal) \right)
    \label{eq:kl_div_corpus}
\end{equation}
The  intuitive interpretation is that we expect the observed label probability $\hat{P}$ on set $X$ to match the prior probability $Q$. Consequently, we should select an ordering that assigns probabilities  labels that are similar to the prior. 

As mentioned in~\cref{sec:fewshot}, if we do not have access to a prior label distribution, we need to assume a uninformative prior and simply set $P(y)=1/|\Ycal|$, i.e., uniform distribution $\text{Unif}(\Ycal)$ over the label space $\Ycal$.

\subsection{Selecting a Performant Ordering}
\label{sec:select}

The set $\Dcal$ of $k$ in-context examples lead to $k!$ possible orderings. Even for small values of $k$, we can end up with a prohibitively large number of orderings to score and rank, e.g., with $8$ examples, we have to consider $8! = 40,320$ permutations.
We propose a simple sample-then-select solution similar to \citet{lu-etal-2022-fantastically}.\footnote{Alternatively, we could seek to parameterize an ordering $\pi$; we leave this extension for future work.} 
We first randomly sample $K$ permutations from all possible $k!$ permutations, then rank them as in~\cref{eq:kl_div_null} and~\cref{eq:kl_div_corpus}:
\begin{equation}
    \pi^{*}=\underset{\pi}{\arg\min} \,\mathcal{L}(\pi)
\end{equation}

We call our method \emph{Probability Distribution Ordering} (\pdo). 
This choice is independent of the \direct and \pmi approaches (which use a given ordering). 
As a result, we can combine \direct and \pmi approaches with \pdo.

\section{Experiments}
\label{sec:experiments}

Our experiments evaluate the effectiveness of the proposed principles and answer the following research questions:
\begin{enumerate}
    \item  Does \pdo improve in-context learning accuracy and reduce variance?
    \item While vanilla confidence calibration methods (such as temperature scaling) require a labeled development set, can \pdo better calibrate a model without a labeled development set?
\end{enumerate}

\subsection{Experimental Setup}
\label{sec:experiment_setup}
We conduct experiments on 13 text classification datasets including binary and multi-label classifications, as well as balanced and imbalanced datasets. 
The details of these datasets and the prompt templates are in \cref{sec:datasets}.
We also use 9 autoregressive language models of varying sizes to demonstrate the robustness of our proposed approach.
Our approach falls into the category of \emph{corpus-level} ICL \cite{wu-etal-2023-self} (or \emph{task-level} ICL) where we select the best-performing template with or without a validation set and then equally apply this template to all test examples during in-context learning. 

As scoring functions and ordering selection are two orthogonal procedures, we experiment with two different scoring approaches---\direct and \pmi---to demonstrate the effectiveness of \pdo. 

\subsection{Baselines}
\label{sec:baselines}

We compare against a number of baseline configurations detailed below. 

\paragraph{Random.} For a given set of in-context examples $\Dcal$, we sample a set of orderings $\{\pi_1(\Dcal), \pi_2(\Dcal), \cdots, \pi_n(\Dcal)\}$, perform in-context learning with \direct or \pmi, and average the performance metrics across the set of orderings. 
We do not perform any ordering selection. This configuration is as an important baseline for \settingA where we do not have an unlabeled set $X$. 

\paragraph{GlobalE and LocalE.} 
Following~\citet{lu-etal-2022-fantastically}, \textbf{GlobalE} gathers the predicted labels of all examples in unlabeled development set $X$, and selects the ordering with the minimum KL-divergence between uniform distribution and predicted label distribution. 
Enforcing a uniform distribution of predicted labels potentially degrades performance on imbalanced datasets. 
\textbf{LocalE} is similar to~\cref{eq:kl_div_corpus}, but instead computes KL-divergence between uniform distribution and the probability distribution of each sample in unlabeled development set $X$:
\begin{equation}
    \mathcal{L}(\pi) = \sum_{x \in X} D_{\text{KL}} \big(P(\mathcal{Y}|\pi,x) \,||\, \text{Unif.}(\mathcal{Y}) \big). 
\end{equation}
This criterion implicitly encourages the language model to predict a uniform distribution over individual samples, whereas~\cref{eq:kl_div_corpus} minimizes the divergence globally.
\textbf{LocalE} and \textbf{GlobalE} serve as important baselines for \settingB and \settingC, where we assume unlabeled development set $X$.

\paragraph{Oracle.}
We select the orderings that lead to the best performance on $X$, assuming access to an oracle that provides ground truth labels. 
This configuration serves as an upper bound for all performant ordering selection approaches.

\paragraph{Combining PDO with Direct and PMI.} 
Finally, for each model-dataset pair, as described in \cref{sec:method}, we combine \pdo with the \direct and \pmi approaches. 
These configurations are referred to as \textbf{PDO-Direct} and \textbf{PDO-PMI}, respectively.  
\cref{tab:classification}, for example, shows the performance of OPT-13B and LLaMA-13B using both \direct and \pmi combined with the three settings of \pdo.

\begin{figure*}[!t]
\begin{subfigure}{0.48\textwidth}
    \includegraphics[width=\linewidth]{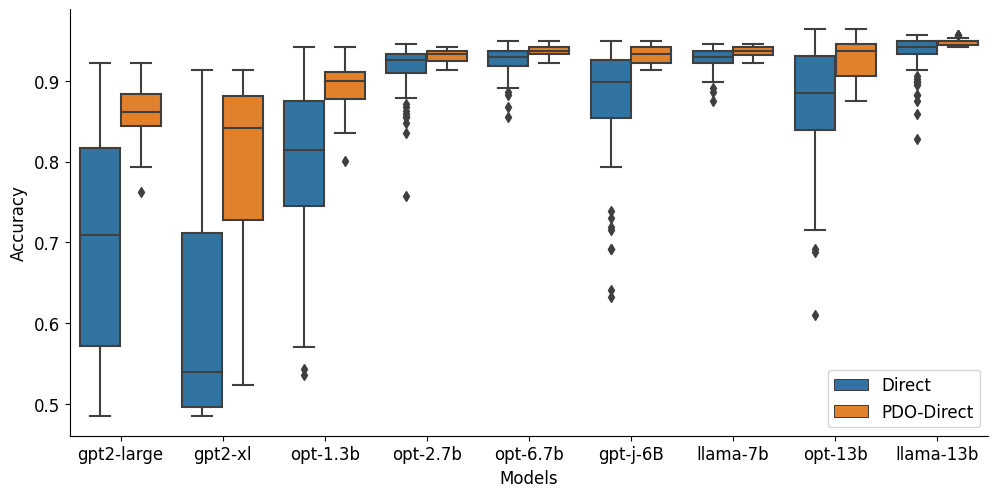}
    \caption{SST-2 results with \direct and \textbf{\pdo-\direct}.}
    \label{fig:sst2-direct}
\end{subfigure}
\hspace{-5pt}
\begin{subfigure}{0.48\textwidth}
    \includegraphics[width=\linewidth]{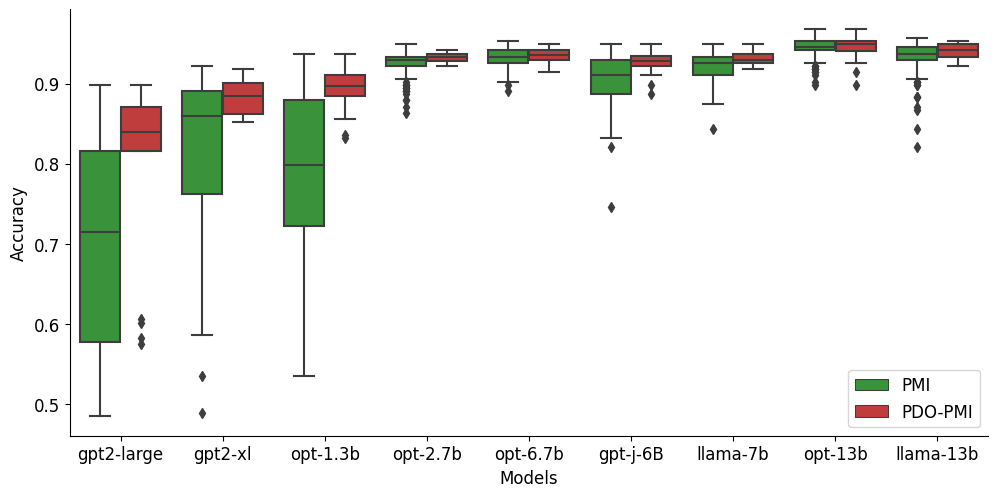}
    \caption{SST-2 results with \pmi and \textbf{\pdo-\pmi}.}
    \label{fig:sst2-pmi}
\end{subfigure}
\caption{
SST-2 results with different language models.  
}
\label{fig:sst2}
\end{figure*}

\begin{figure*}[!t]
\begin{subfigure}{0.48\textwidth}
    \includegraphics[width=\linewidth]{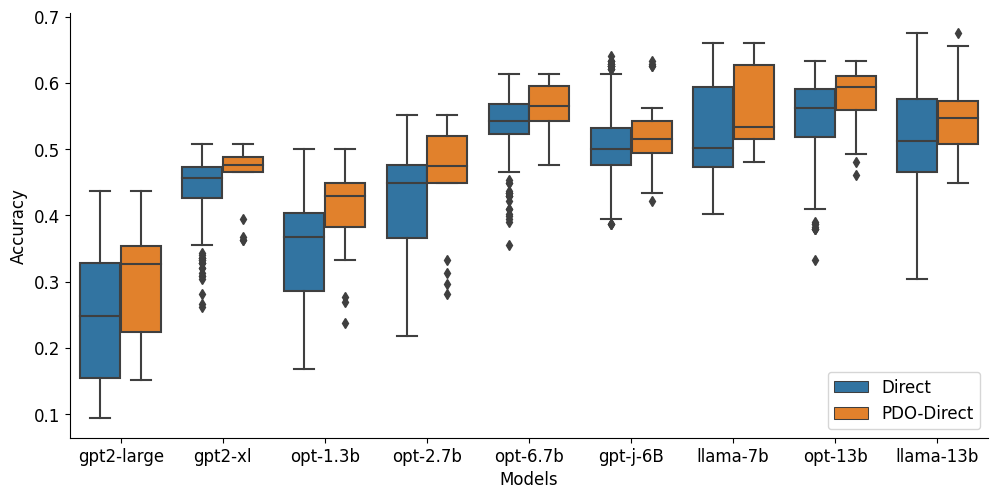}
    \caption{Yahoo Topics results with \direct and \textbf{\pdo-\direct}.}
    \label{fig:yahoo-direct}
\end{subfigure}
\hspace{-5pt}
\begin{subfigure}{0.48\textwidth}
    \includegraphics[width=\linewidth]{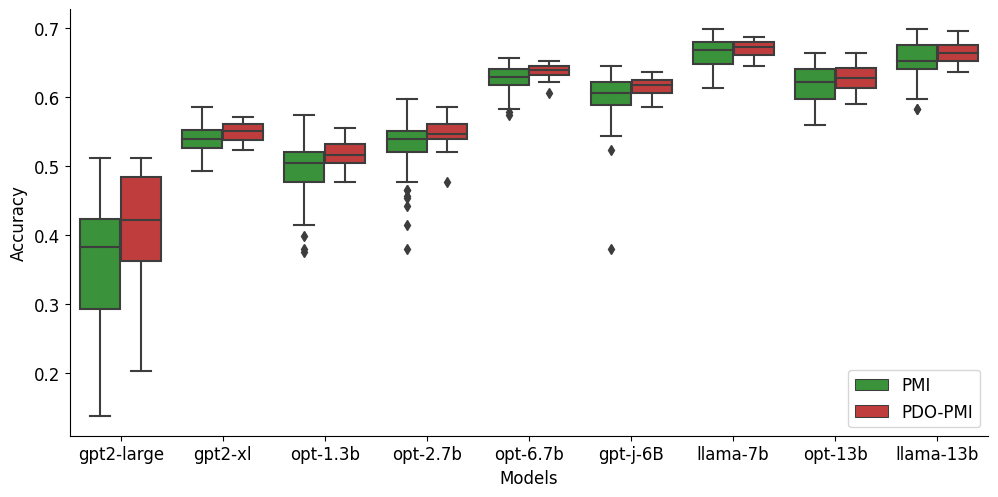}
    \caption{Yahoo Topics results with \pmi and \textbf{\pdo-\pmi}.}
    \label{fig:yahoo-pmi}
\end{subfigure}
\caption{
Yahoo topic results with different language models. 
}
\vspace{-15pt}
\label{fig:yahoo}
\end{figure*}

\subsection{Evaluation}
\label{subsec:evaluation}
For each dataset, we use 8 in-context examples (shots)  and 5 different random seeds (by default), i.e., 5 different sets of uniformly sampled in-context examples. 
For each set of in-context examples, we randomly sample 24 orderings and report the average accuracy and Expected Calibration Error (ECE) \cite{naeini2015obtaining} computed with a fixed number of 100 bins. 
For \textbf{GlobalE}, \textbf{LocalE}, \pdo and \textbf{Oracle}, we select top-4 orderings out of 24 sampled orderings. 
The results of \textbf{Random} are averaged over $24\times5=120$ runs while the results of \textbf{GlobalE}, \textbf{LocalE}, \pdo and \textbf{Oracle} are averaged over $4\times5=20$ runs.
We randomly sample 256 instances from the training set (not overlapping with the 8 in-context examples) to form an unlabeled set $X$, and use the label distribution as the informative prior probability distribution.

We benchmark the performance of various approaches with 9 autoregressive LLMs with 770M to 13B parameters: 
GPT2-large (770M), GPT2-xl (1.5B) \cite{radford2019language}, OPT-1.3B,  OPT-2.7B,  OPT-6.7B,  OPT-13B~\cite{zhang2022opt}, GPT-J-6B \cite{gpt-j}, and the more recent LLaMA-7B and LLaMA-13B~\cite{touvron2023llama}. 
For the rest of this section, we mainly discuss results on OPT-13B and LLaMA-13B; the results on smaller models show similar trends and we show complete results in~\cref{sec:complete-results}.\footnote{
Our code is available at \href{https://github.com/zhichaoxu-shufe/Prompt-ordering-label-distributions}{Link to Github}.}

\subsection{Results and Analysis}
\label{sec:classification-results}

\begin{table}[!t]
\centering
\caption{Accuracy and standard deviation measured in {\%} on OPT-13B and LLaMA-13B respectively, e.g., $65.8_{7.2}$ means a $65.8\%$ accuracy with a $7.2\%$ variance.}
\vspace{0pt}
\resizebox{\columnwidth}{!}{ 
\begin{tabular}{llll}
\toprule
Methods 
& \begin{tabular}[c]{@{}l@{}l} Avg. \\ \, \\ \end{tabular} 
& \begin{tabular}[c]{@{}l@{}l} Avg. \\ Balanced\\ \end{tabular} 
& \begin{tabular}[c]{@{}l@{}l} Avg. \\ Imbalanced \\ \end{tabular} 
\\ 
\rowcolor{lightgray}
\multicolumn{4}{l}{\emph{\textbf{OPT-13B Direct}}}\\
Random & $65.8_{7.2}$  & $80.9_{5.0}$ & $52.7_{9.0}$ \\
LocalE & $65.2_{6.6}$ & $\textbf{81.4}_{4.5}$ & $\textbf{54.2}_{8.4}$ \\
GlobalE & $\textbf{66.1}_{5.5}$ & $80.4_{2.9}$ & $53.8_{7.9}$ \\ 
\hline
PDO (\settingA) & $66.8_{6.5}$  & $82.7_{3.6}$ & ${53.2}_{9.0}$ \\
PDO (\settingB) & $66.8_{6.2}$ & $81.5_{4.1}$ & $54.2_{8.0}$ \\
PDO (\settingC) & $\textbf{69.6}_{5.6}$ & $\textbf{84.2}_{2.9}$ & $\textbf{57.2}_{7.8}$ \\
\hline
Oracle & $70.8_{4.9}$ & $84.6_{2.1}$ & $59.0_{7.3}$ \\
\rowcolor{lightgray}
\multicolumn{4}{l}{\emph{\textbf{OPT-13B PMI}}}\\
Random & $67.0_{8.1}$  & $85.4_{6.9}$ & $51.3_{9.1}$ \\
LocalE & $\textbf{67.8}_{4.5}$ & $\textbf{85.4}_{2.1}$ & $\textbf{52.6}_{6.7}$ \\
GlobalE & $67.2_{3.4}$ & $83.4_{1.4}$ & $53.3_{5.0}$ \\ 
\hline
PDO (\settingA) & $67.8_{7.6}$  & $85.5_{6.9}$ & ${52.7}_{8.1}$ \\
PDO (\settingB) & $69.2_{4.2}$ & $85.5_{2.0}$ & $55.3_{6.1}$ \\
PDO (\settingC) & $\textbf{71.1}_{3.5}$ & $\textbf{86.1}_{1.5}$ & $\textbf{58.3}_{5.3}$ \\
\hline
Oracle & $73.1_{2.8}$ & $87.4_{1.2}$ & $60.9_{4.1}$ \\
\rowcolor{lightgray}
\multicolumn{4}{l}{\emph{\textbf{LLaMA-13B Direct}}}\\
Random & $70.8_{6.1}$ & $83.4_{3.8}$ & $59.8_{8.0}$ \\
LocalE & $70.6_{5.8}$ & $83.2_{4.4}$ & $59.8_{7.0}$ \\
GlobalE & $\textbf{72.9}_{5.0}$ & $\textbf{85.1}_{2.8}$ & $\textbf{62.4}_{6.9}$ \\ 
\hline
PDO (\settingA) & $71.2_{6.0}$ & $83.9_{3.4}$ & $60.3_{8.1}$ \\
PDO (\settingB) & $73.1_{4.8}$ & $\textbf{85.5}_{2.7}$ & $62.5_{5.5}$ \\
PDO (\settingC) & $\textbf{74.0}_{3.3}$ & $85.4_{2.7}$ & $\textbf{64.2}_{3.9}$ \\
\hline
Oracle & $75.5_{3.2}$ & $85.9_{2.5}$ & $66.7_{3.7}$ \\
\rowcolor{lightgray}
\multicolumn{4}{l}{\emph{\textbf{LLaMA-13B PMI}}}\\
Random & $72.2_{6.1}$ & $87.3_{3.8}$ & $59.3_{8.0}$ \\
LocalE & $71.4_{5.0}$ & $86.5_{1.9}$ & $58.5_{7.6}$ \\
GlobalE & $\textbf{73.6}_{3.4}$ & $\textbf{88.0}_{1.1}$ & $\textbf{61.3}_{5.4}$ \\ 
\hline
PDO (\settingA) & $72.0_{6.0}$ & $87.2_{3.4}$ & $59.1_{8.1}$ \\
PDO (\settingB) & $73.9_{3.6}$ & $\textbf{87.9}_{1.4}$ & $61.9_{5.5}$ \\
PDO (\settingC) & $\textbf{75.5}_{2.7}$ & $87.8_{1.4}$ & $\textbf{65.0}_{3.9}$ \\ 
\hline
Oracle & $77.4_{2.1}$ & $89.0_{0.9}$ & $67.4_{3.1}$ \\
\bottomrule
\end{tabular}
}
\label{tab:classification}
\vspace{-10pt}
\end{table}

Increasing model size improves ICL classification performance.
Figures~\ref{fig:sst2} and \ref{fig:yahoo} show the effect of different model choices on predictive accuracy.
As the model size increases, the classification performance also increases whereas the variance decreases. 
This observation is consistent with prior works~\cite{min-etal-2022-rethinking,lu-etal-2022-fantastically}.

In \settingA where no unlabeled set is available, \pdo is competitive (\cref{tab:classification}).
In three out of four sections in \cref{tab:classification}, \pdo outperforms the non-selective baselines, and is only slightly worse than \textbf{Random} with LLaMA-13B \pmi ($72.0\%$ compared to $72.2\%$).

When an unlabeled set is available, but the label prior is unknown (\settingB), \pdo slightly outperforms  \textbf{GlobalE} and \textbf{LocalE}.
For example, for OPT-13B \direct, \pdo achieves on average $66.8\%$ compared to \textbf{LocalE}'s $65.2\%$ and \textbf{GlobalE}'s $66.1\%$, whereas for OPT-13B \pmi, the numbers are $69.2\%$ compared to $67.8\%$ and $67.2\%$. 

\cref{tab:classification} shows that \pdo performance consistently improves with more information. 
For example, the average performance on 13 datasets is $71.2\%$, $73.1\%$ and $74.0\%$ in the \settingA, \settingB, and \settingC settings respectively.
With prior probability distribution known, \pdo outperforms all baselines, and \pdo-\pmi further improves the classification performance

\begin{figure}[t]
\centering
\includegraphics[width=\columnwidth]{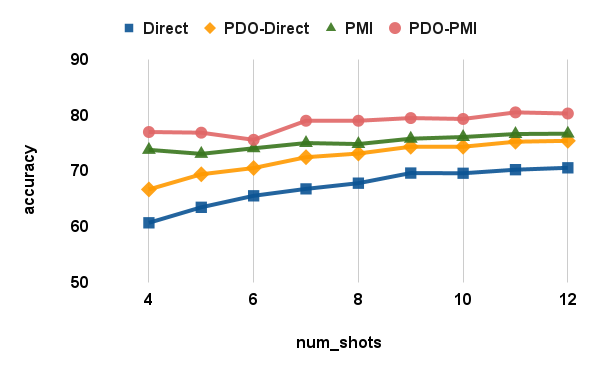}
\vspace{-20pt}
\caption{
We show the mean accuracy over 5 topic classification datasets across different numbers of in-context training examples (from 4 to 12) under \settingC. 
The backbone LLM is LLaMA-7B. 
\pdo's improvement is consistent with different numbers of samples.
}
\label{fig:ablation}
\vspace{-5pt}
\end{figure}

\pdo's performance improvement is consistent across different numbers of in-context examples.
\cref{fig:ablation} shows an ablation study with varying numbers of in-context examples. 
We report mean accuracy on 5 topic classification datasets with LLaMA-7B. 
We observe that as the number of samples increases, the mean accuracy also improves. Further, using \pdo  consistently improves performance over the non-selective baselines.

\pdo reduces in-context learning miscalibration.
From \cref{tab:calibration} we notice that under three different settings, \pdo can all reduce model miscalibration compared to the baselines.
For example, for OPT-13B \direct \settingA, \pdo reports an average ECE result of 17.9\% compared to \textbf{Random}'s 20.4\%; for \settingB, \pdo reports an average of 17.1\% compared to \textbf{LocalE}'s 19.9\% and \textbf{GlobalE}'s 17.3\%.
Notably, \textbf{Oracle} reports 17.4\%, meaning that predictive accuracy and model calibration are not always positively correlated, whereas \pdo can effectively prevent the model's prediction from being too confident.

\begin{table}[!t]
\centering
\caption{Expected Calibration Error results (measured in \%) on OPT-13B and LLaMA-13B, respectively.}
\vspace{0pt}
\resizebox{\columnwidth}{!}{
\begin{tabular}{lrrr}
\toprule
Methods 
& \begin{tabular}[c]{@{}l@{}l} Avg. \\ \, \\ \end{tabular} 
& \begin{tabular}[c]{@{}l@{}l} Avg. \\ Balanced\\ \end{tabular} 
& \begin{tabular}[c]{@{}l@{}l} Avg. \\ Imbalanced \\ \end{tabular} 
\\ 
\rowcolor{lightgray}
\multicolumn{4}{l}{\emph{\textbf{OPT-13B Direct}}}\\
Random & $20.4$  & $13.3$ & $26.4$ \\
LocalE & $19.9$ & $13.4$ & $25.5$ \\
GlobalE & $\textbf{17.3}$ & $\textbf{12.6}$ & $\textbf{21.2}$ \\ 
\hline
PDO (\settingA) & $17.9$  & $13.0$ & $22.2$ \\
PDO (\settingB) & $17.1$ & $13.6$ & $20.2$ \\
PDO (\settingC) & $\textbf{16.6}$ & $\textbf{12.5}$ & $\textbf{20.1}$ \\
\hline
Oracle & $17.4$ & $12.7$ & $21.5$ \\
\rowcolor{lightgray}
\multicolumn{4}{l}{\emph{\textbf{OPT-13B PMI}}}\\
Random & $20.2$  & $18.2$ & $21.8$ \\
LocalE & $18.0$ & $18.2$ & $17.9$ \\
GlobalE & $\textbf{17.7}$ & $\textbf{17.9}$ & $\textbf{17.7}$ \\ 
\hline
PDO (\settingA) & $17.9$  & $\textbf{17.4}$ & $20.7$ \\
PDO (\settingB) & $\textbf{17.2}$ & $18.1$ & $\textbf{16.4}$ \\
PDO (\settingC) & $17.6$ & $\textbf{17.4}$ & $17.7$ \\
\hline
Oracle & $18.6$ & $17.8$ & $19.4$ \\
\rowcolor{lightgray}
\multicolumn{4}{l}{\emph{\textbf{LLaMA-13B Direct}}}\\
Random & $16.1$ & $12.6$ & $19.1$ \\
LocalE & $15.4$ & $13.2$ & $17.4$ \\
GlobalE & $\textbf{14.4}$ & $\textbf{12.2}$ & $\textbf{16.4}$ \\ 
\hline
PDO (\settingA) & $14.8$ & $12.7$ & $16.5$ \\
PDO (\settingB) & $14.2$ & $12.1$ & $16.0$ \\
PDO (\settingC) & $\textbf{13.9}$ & $\textbf{12.0}$ & $\textbf{15.5}$ \\
\hline
Oracle & $15.4$ & $12.0$ & $18.3$ \\
\rowcolor{lightgray}
\multicolumn{4}{l}{\emph{\textbf{LLaMA-13B PMI}}}\\
Random & $17.8$ & $16.9$ & $18.7$ \\
LocalE & $17.4$ & $\textbf{16.5}$ & $18.0$ \\
GlobalE & $\textbf{16.9}$ & $16.8$ & $\textbf{17.0}$ \\ 
\hline
PDO (\settingA) & $17.4$ & $16.0$ & $18.5$ \\
PDO (\settingB) & $16.8$ & $\textbf{15.9}$ & $17.5$ \\
PDO (\settingC) & $\textbf{16.5}$ & $\textbf{15.9}$ & $\textbf{17.2}$ \\ 
\hline
Oracle & $17.6$ & $16.6$ & $18.5$ \\
\bottomrule
\end{tabular}
}
\label{tab:calibration}
\vspace{-5pt}
\end{table}

\subsection{In-context example selection}
\label{subsec:sample_selection}
So far, we have evaluated \pdo for selecting performant orderings.
Can the principles that drive it also be used to select in-context examples at the task-level?
We follow a similar setup as in \cref{subsec:evaluation}, i.e. we randomly sample 120 sets of in-context examples, each set consisting of 8 examples, and we sample one permutation from each set. Then we utilize \pdo to rank them and select the top 20 sets of examples. 
We compare to \text{CondAcc}~\cite{chang-jia-2023-data} for selecting in-context examples at the task level. \text{CondAcc} computes each in-context example's influence on \emph{labeled development set $X^*$}. After determining the top-$k$ influential examples, we place them in the increasing order of influence~\cite{liu-etal-2022-makes}.

From Table~\ref{tab:sample_selection}, we see that in 3 out of 4 sections, \settingA improves performance compared to Random.  \settingB and \settingC consistently outperform Random in all 4 sections. 
\settingC also matches CondAcc's performance despite not using a labeled development set.
These observations show that \pdo's can help select performant samples at the task level.

\begin{table}[!t]
\centering
\caption{Sample selection results (accuracy and standard deviation measured in {\%}) on OPT-13B and LLaMA-13B respectively.}
\vspace{0pt}
\resizebox{\columnwidth}{!}{ 
\begin{tabular}{llll}
\toprule
Methods 
& \begin{tabular}[c]{@{}l@{}l} Avg. \\ \, \\ \end{tabular} 
& \begin{tabular}[c]{@{}l@{}l} Avg. \\ Balanced\\ \end{tabular} 
& \begin{tabular}[c]{@{}l@{}l} Avg. \\ Imbalanced \\ \end{tabular} 
\\ 
\rowcolor{lightgray}
\multicolumn{4}{l}{\emph{\textbf{OPT-13B Direct}}}\\
Random & $65.9_{7.4}$  & $80.8_{5.7}$ & $53.1_{8.8}$ \\ \hline
PDO (\settingA) & $68.3_{5.3}$  & $84.2_{2.3}$ & $54.7_{7.9}$ \\
PDO (\settingB) & $72.3_{4.1}$ & $86.5_{1.6}$ & $60.1_{6.2}$ \\
PDO (\settingC) & $73.0_{2.9}$ & $86.5_{1.5}$ & $61.5_{4.1}$ \\ \hline
CondAcc & $72.7_{0.0}$ & $86.7_{0.0}$ & $61.1_{0.0}$ \\
Oracle & $75.3_{1.5}$ & $87.1_{1.0}$ & $65.1_{1.9}$ \\
\rowcolor{lightgray}
\multicolumn{4}{l}{\emph{\textbf{OPT-13B PMI}}}\\
Random & $67.1_{5.6}$  & $85.2_{2.4}$ & $51.5_{8.3}$ \\ \hline
PDO (\settingA) & $67.7_{5.2}$  & $85.8_{1.9}$ & $52.3_{8.1}$ \\
PDO (\settingB) & $71.6_{3.7}$ & $86.8_{1.4}$ & $58.6_{5.6}$ \\
PDO (\settingC) & $73.1_{3.4}$ & $86.7_{1.5}$ & $61.4_{5.1}$ \\ \hline
CondAcc & $72.9_{0.0}$ & $86.5_{0.0}$ & $61.1_{0.0}$ \\
Oracle & $75.2_{1.7}$ & $88.3_{0.5}$ & $64.1_{2.7}$ \\
\rowcolor{lightgray}
\multicolumn{4}{l}{\emph{\textbf{LLaMA-13B Direct}}}\\
Random & $70.2_{6.7}$ & $82.4_{4.5}$ & $59.6_{8.5}$ \\ \hline
PDO (\settingA) & $71.3_{5.6}$ & $84.3_{2.8}$ & $60.2_{8.1}$ \\
PDO (\settingB) & $73.6_{4.5}$ & $87.4_{1.2}$ & $61.8_{7.3}$ \\
PDO (\settingC) & $75.4_{2.1}$ & $87.2_{1.3}$ & $65.4_{2.9}$ \\ \hline
CondAcc & $75.0_{0.0}$ & $86.5_{0.0}$ & $65.5_{0.0}$ \\
Oracle & $77.6_{1.4}$ & $87.8_{0.9}$ & $68.9_{1.8}$ \\
\rowcolor{lightgray}
\multicolumn{4}{l}{\emph{\textbf{LLaMA-13B PMI}}}\\
Random & $72.1_{5.4}$ & $87.1_{2.1}$ & $59.3_{8.1}$ \\ \hline
PDO (\settingA) & $72.0_{5.5}$ & $87.0_{2.2}$ & $59.1_{8.2}$ \\
PDO (\settingB) & $74.1_{3.3}$ & $88.3_{1.2}$ & $61.8_{5.1}$ \\
PDO (\settingC) & $76.1_{2.4}$ & $88.1_{1.2}$ & $65.7_{3.4}$ \\ \hline
CondAcc & $75.8_{0.0}$ & $87.5_{0.0}$ & $65.6_{0.0}$ \\
Oracle & $78.8_{1.2}$ & $89.7_{0.5}$ & $69.4_{1.8}$ \\
\bottomrule
\end{tabular}
}
\label{tab:sample_selection}
\vspace{-10pt}
\end{table}

\section{Related Work}
\label{sec:related}
\vspace{-5pt}

\citet{brown2020language} first demonstrated that autoregressive LLMs are able to ``learn in context''.
Various strategies have since been proposed to improve in-context learning performance. 
\citet{wei2022chain,kojima2022large} showed that chain-of-thought prompting can improve LLM's performance on reasoning tasks.
A different line of works~\cite{su2022selective,liu-etal-2022-makes,gao2020making,lyu-etal-2023-z} proposes to augment ICL performance via examples/evidence retrieval.

Existing works on prompt engineering for few-shot ICL can be broadly divided into the following directions: 
(1) sample selection~\cite{su2022selective,gao2020making};  
(2) scoring functions, to propose alternative scoring functions to replace raw probability; examples include PMI~\cite{holtzman-etal-2021-surface}, Noisy Channel Classification~\cite{min-etal-2022-noisy}, Contextual Calibration~\cite{zhao2021calibrate};
(3) prompt instruction/order selection \cite{lu-etal-2022-fantastically,wu-etal-2023-self,sorensen-etal-2022-information}.

Prior works have noticed that ICL performance is sensitive to sample choices and ordering. 
\citet{zhao2021calibrate} noticed that models can be biased (recency bias, majority bias) towards certain labels from in-context examples.
\citet{min-etal-2022-rethinking} conducted an empirical study to discuss factors important to ICL performance. 
\citet{lu-etal-2022-fantastically} pointed out ICL performance can vary from near full fine-tuning to random across different orderings of the same set of examples. 
\citet{wu-etal-2023-self} combined retrieval augmentation and leveraged the backbone LLM's confidence to rank and select performant orderings.
Our work closely follows~\citet{lu-etal-2022-fantastically}'s approach and generalizes to different resource settings.

Per~\citet{wu-etal-2023-self}, our method can be categorized to \emph{corpus-level} approaches, i.e., selecting a universal in-context example ordering for all instances.
There are in fact \emph{instance-level} approaches, i.e., selecting performant orderings for each single test instance~\cite{su2022selective,liu-etal-2022-makes}. 
Unlike task-level selection/ordering, instance-level approaches require more computational effort because the selection/ordering needs to be performed for every instance.

Limited works have discussed the confidence calibration problem in ICL. 
As argued by prior works~\cite{guo2017calibration,niculescu2005predicting,platt1999probabilistic}, reliable confidence measurement is critical in classification problems, especially high-risk decision scenarios. 
Existing confidence calibration methods including temperature scaling and Platt scaling mostly require labeled samples to tune hyperparameters while our method do not require labeled samples.
Our work can serve as a motivation for future works to add confidence calibration as an evaluation metric for ICL.

Prior works~\cite{yu2014learning,dulac2019deep} on \emph{Learning from Label Proportions} (LLP) focus on the settings where an instance-level labeling is unavailable. 
Considering we have $N$ bags, each consisting of $n_i$ examples, we do not have access to each corresponding label, but instead, we have access to the label proportions of each bag.
In this case, we can still learn a classifier using the label proportions of all $N$ bags as weak supervision signals (see theoretical proof in~\citet{yu2014learning,zhang2022learning}).
\citet{dulac2019deep} discussed choices of empirical loss functions for image classification task and found a classifier could be learned by minimizing KL divergence between predicted label proportions and true label proportions.

\section{Conclusions}
\label{sec:conclusions}
\vspace{-5pt}

In this paper, we aim to optimize in-context example ordering to improve ICL performance.
We rigorously examine three problem settings based on the availability of labeled examples and propose two principles in selecting performant orderings.
Our approach, referred to as the Probability Distribution Ordering (PDO), leverages the model's output probability distributions.
Via extensive experiments, we demonstrate that our approach requires a trivial amount of extra computation and outperforms the baselines by improving classification accuracy and reducing   model miscalibration.

\section{Limitations}
\label{sec:limitation}

Due to limited bandwidth and budget, we only experiment with autoregressive LLMs no larger than 13B. 
The effectiveness with encoder-decoder models such as those from T5 family~\cite{raffel2020exploring} are not studied.
Examining our findings on commercial language models such as GPT-4 requires further experiments.

The proposed principles require computing the output probability distributions, thus they are not trivial to extend to generation tasks such as Open QA and summarization.
We believe a potential future direction is to generalize the proposed approach to natural language generation tasks.

\section*{Acknowledgements}
We would thank members of UtahNLP for their constructive feedback.
This material is based upon work supported by NSF under grants \#2134223, \#2205418, \#2007398, \#2217154, \#1822877 and \#1801446. The support and resources from the Center for High Performance Computing at the University of Utah are gratefully acknowledged.



\bibliography{anthology}

\appendix
\label{sec:appendix}

\section{Details on Datasets and Templates}
\label{sec:datasets}

We provide details on the 13 datasets and templates in~\cref{tab:dataset} and~\cref{tab:templates}, respectively. 
All datasets licenses are available for public use.
\begin{table}[!ht]
\centering
\caption{Details on datasets.}
\resizebox{\columnwidth}{!}{
\begin{tabular}{lrr}
\toprule
Dataset & \# Classes & Balanced \\
\midrule
\multicolumn{3}{l}{\emph{Sentiment Classification}} \\
SST2~\cite{socher-etal-2013-recursive} & 2 & \cmark \\
SST5~\cite{socher-etal-2013-recursive} & 5 & \xmark \\
CR~\cite{hu2004mining} & 2 & \cmark \\
MR~\cite{pang-lee-2005-seeing} & 2 & \cmark \\
financial\_phrasebank~\cite{malo2014good} & 3 & \xmark \\ 
\midrule
\multicolumn{3}{l}{\emph{Topic Classification}} \\
AG News~\cite{zhang2015character} & 4 & \cmark \\
TREC~\cite{voorhees2000building} & 6 & \cmark \\
Yahoo Topics~\cite{zhang2015character} & 10 & \xmark \\
Dbpedia~\cite{lehmann2015dbpedia} & 14 & \xmark \\
Subj~\cite{pang-lee-2005-seeing} & 2 & \xmark \\
\midrule
\multicolumn{3}{l}{\emph{Toxicity Detection}} \\
Tweet Offensive~\cite{barbieri-etal-2020-tweeteval} & 2 & \xmark \\
Tweet Irony~\cite{barbieri-etal-2020-tweeteval} & 2 & \xmark \\
Tweet Hate~\cite{barbieri-etal-2020-tweeteval} & 2 & \xmark \\
\bottomrule
\end{tabular}
}
\vspace{0pt}
\label{tab:dataset}
\end{table}

\begin{table*}[!ht]
\vspace{0pt}
\centering
\caption{Templates and label tokens. We use minimum templates and single token labels similar to \cite{lu-etal-2022-fantastically,wu2022self}.}
\resizebox{1.0\textwidth}{!}{
\begin{tabular}{lll}
\toprule
Dataset & Template & Label Tokens \\ 
\midrule
SST2, CR, MR & \textcolor{MidnightBlue}{Review:} [INPUT]\textcolor{MidnightBlue}{\textbackslash nSentiment:} [LABEL] & positive, negative \\
\hline
SST5 & \textcolor{MidnightBlue}{Review:} [INPUT]\textcolor{MidnightBlue}{\textbackslash nSentiment:} [LABEL] & terrible, bad, okay, good, great \\
\hline
financial\_phrasebank & \textcolor{MidnightBlue}{News:} [INPUT]\textcolor{MidnightBlue}{\textbackslash nSentiment:} [LABEL] & positive, negative \\
\hline
Subj & \textcolor{MidnightBlue}{Input:} [INPUT]\textcolor{MidnightBlue}{\textbackslash nType:} [LABEL] & subjective, objective \\
\hline
AG News & \textcolor{MidnightBlue}{Input:} [INPUT]\textcolor{MidnightBlue}{\textbackslash nType:} [LABEL] & sports, business, world, technology \\
\hline
TREC & \textcolor{MidnightBlue}{Question:} [INPUT]\textcolor{MidnightBlue}{\textbackslash nType:} [LABEL] & \begin{tabular}[c]{@{}l@{}l} description, entity, expression\\location, number, human  \end{tabular} \\
\hline
Dbpedia & \textcolor{MidnightBlue}{Input:} [INPUT]\textcolor{MidnightBlue}{\textbackslash nType:} [LABEL] & \begin{tabular}[c]{@{}l@{}l@{}l@{}l} company, school, artist, athlete\\politics, transportation, building\\nature, village, animal, plant\\ album, film, book  \end{tabular} \\
\hline
Yahoo Topics & \textcolor{MidnightBlue}{Question:} [INPUT]\textcolor{MidnightBlue}{\textbackslash nTopic:} [LABEL] & \begin{tabular}[c]{@{}l@{}l@{}l} culture, science, health, politics\\education, electronics, entertainment\\business, sports, relationship\end{tabular} \\
\hline
Tweet Irony & \textcolor{MidnightBlue}{Tweet:} [INPUT]\textcolor{MidnightBlue}{\textbackslash nLabel:} [LABEL] & ironic, neutral \\ 
\hline
Tweet Hate & \textcolor{MidnightBlue}{Tweet:} [INPUT]\textcolor{MidnightBlue}{\textbackslash nLabel:} [LABEL] & hate, neutral \\ 
\hline
Tweet Offensive & \textcolor{MidnightBlue}{Tweet:} [INPUT]\textcolor{MidnightBlue}{\textbackslash nLabel:} [LABEL] & offensive, neutral \\ 

\bottomrule
\end{tabular}
}
\vspace{0pt}
\label{tab:templates}
\end{table*}

\section{Computational Complexity of PDO}
For a fixed set of $K$ in-context examples, we sample up to $k$ permutations from $K$! possible permutations. The random baseline does not incur any additional computation cost.

In \settingA (no unlabeled development set available), we perform in total $k$ forward passes to select the permutations that are less biased towards certain labels.

For \settingB and \settingC, for each permutation, we perform forward passes on all instances in the development set $X$, therefore requiring in total $k \times |X|$ forward passes, where $|X|$ denotes the size of the unlabeled development set. 
The computational cost for \settingB and \settingC is the same as GlobalE and LocalE baselines~\cite{lu-etal-2022-fantastically}.

\section{Extended Related Work}
\label{sec:extended_related_work}
Existing works on selecting in-context examples can be categorized into two classes: (i) instance-level example selection, i.e. to select a set of in-context examples (and its ordering) for each instance in the test set, and (ii) corpus-level/task-level example selection, i.e. to select a set of high quality in-context examples and apply them equality to all test instances. 
\citet{wu-etal-2023-self} argue that instance-level example selection can achieve high performance. 
On the other hand, instance-level example selection incurs additional computational costs at the inference time, and its performance suffers from potential degradation when the size of high-quality annotated examples is small.
Only a few prior works focus on task-level example selection. 
For example, \citet{chang-jia-2023-data} propose to utilize influence functions~\cite{koh2017understanding} to calculate the influence score for each individual instance in the training set, and select the most influential ones as in-context examples. 
Their experiment results show that by carefully selecting in-context examples with high influence scores, the performance of in-context learning can be improved while the variance can be reduced.
A concurrent work~\cite{nguyen2023context} show that increasing the number of influential examples can further improve performance. 
VoteK~\cite{su2022selective} can be seen as a special combination of instance-level and task-level ICL, where first in the task-level, a relatively large set of unlabeled instances is selected to be annotated, then at the inference time, for each test instance, a specific small set of annotated examples are selected as in-context examples (instance level).
In \cref{subsec:sample_selection} we show that PDO can achieve comparable performance to a task-level example selection method CondAcc~\cite{chang-jia-2023-data}, while requiring no labeled development set $X^*$ to compute the influence scores.

\section{Complete Results}
\label{sec:complete-results}

\begin{table*}[t]
\centering
\caption{Complete classification results (measured by accuracy), Part 1.}
\vspace{0pt}
\resizebox{\textwidth}{!}{
\begin{tabular}{llrrrrrrrrrrrr}
\toprule
\, & SST2 & CR & MR & SUBJ & SST5 & AGNews & TREC & \begin{tabular}[c]{@{}l@{}l} Yahoo\\ Topics \end{tabular} & Dbpedia & FPB & \begin{tabular}[c]{@{}l@{}l} Tweet\\Offensive \end{tabular} & \begin{tabular}[c]{@{}l@{}l} Tweet\\ Irony \end{tabular} & \begin{tabular}[c]{@{}l@{}l} Tweet\\ Hate \end{tabular}
\\ 
\rowcolor{lightgray}
\multicolumn{14}{l}{\emph{\textbf{GPT2-large Direct}}}\\
Random & 0.698 & 0.606 & 0.690 & 0.541 & 0.332 & 0.491 & 0.386 & 0.247 & 0.444 & 0.461 & 0.503 & 0.481 & 0.461 \\
LocalE & 0.798 & 0.785 & 0.837 & 0.656 & 0.369 & 0.550 & 0.402 & 0.260 & 0.421 & 0.595 & 0.524 & 0.501 & 0.478 \\
GlobalE & 0.812 & 0.784 & 0.821 & 0.632 & 0.376 & 0.558 & 0.423 & 0.275 & 0.369 & 0.544 & 0.475 & 0.492 & 0.484 \\ \hline
PDO (\settingA) & 0.677 & 0.754 & 0.721 & 0.555 & 0.364 & 0.525 & 0.394 & 0.275 & 0.475 & 0.562 & 0.458 & 0.497 & 0.479 \\
PDO (\settingB) & 0.797 & 0.784 & 0.837 & 0.655 & 0.369 & 0.550 & 0.439 & 0.263 & 0.421 & 0.594 & 0.522 & 0.500 & 0.477 \\
PDO (\settingC) & 0.859 & 0.727 & 0.841 & 0.666 & 0.381 & 0.609 & 0.439 & 0.291 & 0.565 & 0.618 & 0.624 & 0.508 & 0.497 \\ \hline
Oracle & 0.867 & 0.794 & 0.854 & 0.670 & 0.429 & 0.618 & 0.456 & 0.317 & 0.575 & 0.647 & 0.637 & 0.528 & 0.511 \\

\rowcolor{lightgray}
\multicolumn{14}{l}{\emph{\textbf{GPT2-large PMI}}}\\
Random & 0.694 & 0.727 & 0.718 & 0.623 & 0.375 & 0.658 & 0.423 & 0.362 & 0.766 & 0.605 & 0.419 & 0.500 & 0.508 \\
LocalE & 0.768 & 0.733 & 0.789 & 0.735 & 0.425 & 0.659 & 0.464 & 0.364 & 0.709 & 0.637 & 0.493 & 0.495 & 0.509 \\
GlobalE & 0.755 & 0.712 & 0.799 & 0.735 & 0.411 & 0.652 & 0.492 & 0.370 & 0.685 & 0.612 & 0.501 & 0.500 & 0.475 \\ 
\hline
PDO (\settingA) & 0.658 & 0.796 & 0.723 & 0.599 & 0.386 & 0.627 & 0.430 & 0.372 & 0.763 & 0.635 & 0.408 & 0.504 & 0.516 \\ 
PDO (\settingB) & 0.769 & 0.733 & 0.790 & 0.730 & 0.425 & 0.660 & 0.488 & 0.364 & 0.709 & 0.637 & 0.494 & 0.495 & 0.513 \\
PDO (\settingC) & 0.801 & 0.811 & 0.832 & 0.746 & 0.412 & 0.730 & 0.488 & 0.407 & 0.812 & 0.721 & 0.584 & 0.499 & 0.556 \\
\hline
Oracle & 0.811 & 0.831 & 0.851 & 0.766 & 0.457 & 0.752 & 0.517 & 0.436 & 0.829 & 0.775 & 0.589 & 0.540 & 0.581 \\

\rowcolor{lightgray}
\multicolumn{14}{l}{\emph{\textbf{GPT2-xl Direct}}}\\
Random & 0.603 & 0.576 & 0.582 & 0.600 & 0.352 & 0.674 & 0.425 & 0.434 & 0.706 & 0.483 & 0.404 & 0.515 & 0.434 \\
LocalE & 0.727 & 0.661 & 0.671 & 0.711 & 0.371 & 0.660 & 0.415 & 0.430 & 0.673 & 0.555 & 0.529 & 0.504 & 0.466 \\
GlobalE & 0.719 & 0.671 & 0.705 & 0.689 & 0.362 & 0.656 & 0.422 & 0.423 & 0.671 & 0.533 & 0.489 & 0.495 & 0.463 \\
\hline
PDO (\settingA) & 0.752 & 0.658 & 0.647 & 0.653 & 0.361 & 0.719 & 0.441 & 0.447 & 0.715 & 0.468 & 0.429 & 0.491 & 0.431 \\
PDO (\settingB) & 0.726 & 0.661 & 0.672 & 0.710 & 0.372 & 0.660 & 0.452 & 0.432 & 0.673 & 0.554 & 0.529 & 0.504 & 0.466 \\
PDO (\settingC) & 0.788 & 0.683 & 0.704 & 0.708 & 0.425 & 0.759 & 0.452 & 0.457 & 0.769 & 0.638 & 0.592 & 0.517 & 0.495 \\
\hline
Oracle & 0.790 & 0.700 & 0.714 & 0.733 & 0.464 & 0.768 & 0.489 & 0.464 & 0.778 & 0.650 & 0.595 & 0.545 & 0.515 \\
\rowcolor{lightgray}
\multicolumn{14}{l}{\emph{\textbf{GPT2-xl PMI}}}\\
Random & 0.818 & 0.801 & 0.770 & 0.617 & 0.267 & 0.799 & 0.473 & 0.539 & 0.818 & 0.441 & 0.414 & 0.468 & 0.451 \\
LocalE & 0.828 & 0.813 & 0.784 & 0.718 & 0.318 & 0.764 & 0.480 & 0.542 & 0.804 & 0.526 & 0.483 & 0.469 & 0.461 \\
GlobalE & 0.811 & 0.825 & 0.792 & 0.733 & 0.285 & 0.766 & 0.505 & 0.530 & 0.801 & 0.519 & 0.490 & 0.441 & 0.460 \\
\hline
PDO (\settingA) & 0.798 & 0.780 & 0.736 & 0.654 & 0.282 & 0.805 & 0.500 & 0.539 & 0.833 & 0.479 & 0.429 & 0.474 & 0.429 \\
PDO (\settingB) & 0.825 & 0.816 & 0.786 & 0.719 & 0.318 & 0.763 & 0.523 & 0.543 & 0.804 & 0.525 & 0.482 & 0.464 & 0.463 \\
PDO (\settingC) & 0.884 & 0.878 & 0.864 & 0.726 & 0.327 & 0.825 & 0.523 & 0.551 & 0.840 & 0.555 & 0.515 & 0.481 & 0.534 \\
\hline
Oracle & 0.896 & 0.887 & 0.867 & 0.733 & 0.345 & 0.839 & 0.551 & 0.562 & 0.847 & 0.561 & 0.533 & 0.505 & 0.541 \\

\rowcolor{lightgray}
\multicolumn{14}{l}{\emph{\textbf{OPT-1.3B Direct}}}\\
Random & 0.800 & 0.879 & 0.794 & 0.600 & 0.404 & 0.725 & 0.384 & 0.348 & 0.809 & 0.686 & 0.504 & 0.471 & 0.535 \\
LocalE & 0.829 & 0.902 & 0.833 & 0.682 & 0.391 & 0.732 & 0.405 & 0.360 & 0.791 & 0.707 & 0.554 & 0.481 & 0.488 \\
GlobalE & 0.828 & 0.894 & 0.821 & 0.687 & 0.395 & 0.727 & 0.409 & 0.355 & 0.745 & 0.705 & 0.511 & 0.480 & 0.471 \\
\hline
PDO (\settingA) & 0.826 & 0.899 & 0.846 & 0.585 & 0.400 & 0.729 & 0.378 & 0.360 & 0.810 & 0.713 & 0.533 & 0.478 & 0.528 \\
PDO (\settingB) & 0.830 & 0.902 & 0.835 & 0.682 & 0.391 & 0.731 & 0.419 & 0.361 & 0.791 & 0.706 & 0.553 & 0.484 & 0.491 \\
PDO (\settingC) & 0.891 & 0.913 & 0.879 & 0.699 & 0.427 & 0.785 & 0.419 & 0.407 & 0.830 & 0.714 & 0.667 & 0.487 & 0.566 \\
\hline
Oracle & 0.901 & 0.918 & 0.883 & 0.721 & 0.479 & 0.791 & 0.455 & 0.409 & 0.835 & 0.756 & 0.683 & 0.506 & 0.590 \\

\rowcolor{lightgray}
\multicolumn{14}{l}{\emph{\textbf{OPT-1.3B PMI}}}\\
Random & 0.791 & 0.907 & 0.868 & 0.576 & 0.364 & 0.752 & 0.418 & 0.496 & 0.871 & 0.639 & 0.523 & 0.504 & 0.497 \\
LocalE & 0.825 & 0.897 & 0.846 & 0.699 & 0.385 & 0.776 & 0.419 & 0.478 & 0.872 & 0.701 & 0.520 & 0.492 & 0.481 \\
GlobalE & 0.805 & 0.899 & 0.852 & 0.695 & 0.390 & 0.766 & 0.453 & 0.470 & 0.868 & 0.701 & 0.505 & 0.497 & 0.481 \\
\hline
PDO (\settingA) & 0.825 & 0.904 & 0.855 & 0.588 & 0.330 & 0.764 & 0.425 & 0.482 & 0.873 & 0.699 & 0.498 & 0.501 & 0.496 \\
PDO (\settingB) & 0.811 & 0.897 & 0.846 & 0.699 & 0.382 & 0.774 & 0.463 & 0.479 & 0.872 & 0.700 & 0.513 & 0.500 & 0.486 \\
PDO (\settingC) & 0.896 & 0.901 & 0.882 & 0.706 & 0.428 & 0.791 & 0.463 & 0.519 & 0.882 & 0.695 & 0.676 & 0.519 & 0.567 \\
\hline
Oracle & 0.901 & 0.927 & 0.897 & 0.723 & 0.442 & 0.817 & 0.513 & 0.535 & 0.890 & 0.745 & 0.682 & 0.547 & 0.592 \\
\bottomrule
\end{tabular}
}
\label{tab:complete_classification_p1}
\vspace{0pt}
\end{table*}

\begin{table*}[t]
\centering
\caption{Complete classification results (measured by accuracy), Part 2.}
\vspace{0pt}
\resizebox{\textwidth}{!}{
\begin{tabular}{llrrrrrrrrrrrr}
\toprule
\, & SST2 & CR & MR & SUBJ & SST5 & AGNews & TREC & \begin{tabular}[c]{@{}l@{}l} Yahoo\\ Topics \end{tabular} & Dbpedia & FPB & \begin{tabular}[c]{@{}l@{}l} Tweet\\Offensive \end{tabular} & \begin{tabular}[c]{@{}l@{}l} Tweet\\ Irony \end{tabular} & \begin{tabular}[c]{@{}l@{}l} Tweet\\ Hate \end{tabular}
\\ 
\rowcolor{lightgray}
\multicolumn{14}{l}{\emph{\textbf{OPT-2.7B Direct}}}\\
Random & 0.917 & 0.895 & 0.891 & 0.653 & 0.455 & 0.747 & 0.418 & 0.420 & 0.844 & 0.626 & 0.563 & 0.537 & 0.516 \\
LocalE & 0.922 & 0.905 & 0.903 & 0.783 & 0.449 & 0.767 & 0.432 & 0.410 & 0.823 & 0.719 & 0.545 & 0.542 & 0.543 \\
GlobalE & 0.929 & 0.925 & 0.919 & 0.803 & 0.475 & 0.811 & 0.457 & 0.461 & 0.859 & 0.791 & 0.564 & 0.539 & 0.567 \\
\hline
PDO (\settingA) & 0.920 & 0.919 & 0.899 & 0.710 & 0.479 & 0.725 & 0.418 & 0.433 & 0.840 & 0.551 & 0.561 & 0.539 & 0.539 \\
PDO (\settingB) & 0.922 & 0.905 & 0.903 & 0.782 & 0.449 & 0.768 & 0.467 & 0.407 & 0.823 & 0.719 & 0.546 & 0.539 & 0.544 \\
PDO (\settingC) & 0.929 & 0.925 & 0.917 & 0.800 & 0.493 & 0.816 & 0.467 & 0.458 & 0.856 & 0.786 & 0.673 & 0.547 & 0.588 \\
\hline
Oracle & 0.937 & 0.929 & 0.921 & 0.809 & 0.508 & 0.818 & 0.488 & 0.464 & 0.864 & 0.834 & 0.684 & 0.553 & 0.611 \\

\rowcolor{lightgray}
\multicolumn{14}{l}{\emph{\textbf{OPT-2.7B PMI}}}\\
Random & 0.925 & 0.921 & 0.902 & 0.693 & 0.399 & 0.728 & 0.409 & 0.532 & 0.865 & 0.567 & 0.564 & 0.532 & 0.547 \\
LocalE & 0.927 & 0.918 & 0.906 & 0.760 & 0.448 & 0.778 & 0.431 & 0.523 & 0.855 & 0.669 & 0.585 & 0.550 & 0.547 \\
GlobalE & 0.912 & 0.911 & 0.922 & 0.751 & 0.433 & 0.761 & 0.442 & 0.52 & 0.851 & 0.667 & 0.571 & 0.551 & 0.540 \\
\hline
PDO (\settingA) & 0.917 & 0.922 & 0.900 & 0.715 & 0.416 & 0.776 & 0.394 & 0.543 & 0.877 & 0.548 & 0.575 & 0.539 & 0.555 \\
PDO (\settingB) & 0.927 & 0.918 & 0.907 & 0.759 & 0.446 & 0.779 & 0.447 & 0.522 & 0.855 & 0.679 & 0.585 & 0.548 & 0.547 \\
PDO (\settingC) & 0.932 & 0.927 & 0.911 & 0.789 & 0.462 & 0.790 & 0.447 & 0.546 & 0.892 & 0.741 & 0.659 & 0.558 & 0.605 \\
\hline
Oracle & 0.939 & 0.937 & 0.922 & 0.815 & 0.475 & 0.803 & 0.521 & 0.563 & 0.898 & 0.762 & 0.671 & 0.573 & 0.636 \\

\rowcolor{lightgray}
\multicolumn{14}{l}{\emph{\textbf{OPT-6.7B Direct}}}\\
Random & 0.924 & 0.871 & 0.911 & 0.690 & 0.450 & 0.703 & 0.448 & 0.532 & 0.868 & 0.742 & 0.620 & 0.525 & 0.501 \\
LocalE & 0.927 & 0.867 & 0.909 & 0.773 & 0.450 & 0.697 & 0.439 & 0.530 & 0.855 & 0.784 & 0.622 & 0.531 & 0.521 \\
GlobalE & 0.938 & 0.915 & 0.915 & 0.788 & 0.452 & 0.771 & 0.437 & 0.557 & 0.878 & 0.768 & 0.613 & 0.538 & 0.533 \\
\hline
PDO (\settingA) & 0.927 & 0.895 & 0.911 & 0.637 & 0.468 & 0.703 & 0.464 & 0.549 & 0.872 & 0.750 & 0.557 & 0.514 & 0.480 \\
PDO (\settingB) & 0.928 & 0.867 & 0.909 & 0.773 & 0.450 & 0.697 & 0.467 & 0.532 & 0.855 & 0.784 & 0.623 & 0.530 & 0.520 \\
PDO (\settingC) & 0.938 & 0.911 & 0.921 & 0.786 & 0.482 & 0.769 & 0.467 & 0.561 & 0.882 & 0.780 & 0.682 & 0.538 & 0.568 \\
\hline
Oracle & 0.942 & 0.917 & 0.927 & 0.808 & 0.516 & 0.773 & 0.517 & 0.565 & 0.886 & 0.832 & 0.687 & 0.546 & 0.583 \\

\rowcolor{lightgray}
\multicolumn{14}{l}{\emph{\textbf{OPT-6.7B PMI}}}\\
Random & 0.932 & 0.897 & 0.914 & 0.638 & 0.415 & 0.823 & 0.418 & 0.627 & 0.874 & 0.760 & 0.459 & 0.504 & 0.489 \\
LocalE & 0.927 & 0.897 & 0.913 & 0.749 & 0.433 & 0.830 & 0.455 & 0.627 & 0.870 & 0.807 & 0.602 & 0.540 & 0.526 \\
GlobalE & 0.932 & 0.875 & 0.901 & 0.744 & 0.432 & 0.811 & 0.471 & 0.620 & 0.861 & 0.801 & 0.591 & 0.525 & 0.512 \\
\hline
PDO (\settingA) & 0.929 & 0.901 & 0.910 & 0.644 & 0.419 & 0.804 & 0.451 & 0.630 & 0.880 & 0.795 & 0.434 & 0.510 & 0.450 \\
PDO (\settingB) & 0.927 & 0.896 & 0.914 & 0.749 & 0.433 & 0.830 & 0.477 & 0.626 & 0.870 & 0.808 & 0.603 & 0.535 & 0.531 \\
PDO (\settingC) & 0.935 & 0.912 & 0.918 & 0.758 & 0.448 & 0.853 & 0.477 & 0.638 & 0.889 & 0.845 & 0.647 & 0.553 & 0.584 \\
\hline
Oracle & 0.944 & 0.919 & 0.929 & 0.768 & 0.465 & 0.870 & 0.513 & 0.646 & 0.896 & 0.869 & 0.651 & 0.565 & 0.597 \\

\rowcolor{lightgray}
\multicolumn{14}{l}{\emph{\textbf{GPT-J-6B Direct}}}\\
Random & 0.879 & 0.831 & 0.876 & 0.728 & 0.447 & 0.795 & 0.498 & 0.509 & 0.851 & 0.535 & 0.563 & 0.486 & 0.469 \\
LocalE & 0.892 & 0.842 & 0.880 & 0.784 & 0.445 & 0.804 & 0.507 & 0.510 & 0.835 & 0.516 & 0.607 & 0.522 & 0.485 \\
GlobalE & 0.928 & 0.883 & 0.905 & 0.794 & 0.438 & 0.813 & 0.536 & 0.524 & 0.866 & 0.505 & 0.593 & 0.521 & 0.494 \\
\hline
PDO (\settingA) & 0.922 & 0.842 & 0.902 & 0.775 & 0.460 & 0.800 & 0.485 & 0.511 & 0.860 & 0.497 & 0.488 & 0.520 & 0.457 \\
PDO (\settingB) & 0.892 & 0.842 & 0.880 & 0.785 & 0.445 & 0.805 & 0.534 & 0.513 & 0.836 & 0.516 & 0.606 & 0.523 & 0.484 \\
PDO (\settingC) & 0.931 & 0.890 & 0.903 & 0.799 & 0.472 & 0.816 & 0.534 & 0.525 & 0.863 & 0.577 & 0.684 & 0.522 & 0.501 \\
\hline
Oracle & 0.935 & 0.893 & 0.908 & 0.823 & 0.489 & 0.820 & 0.548 & 0.533 & 0.871 & 0.607 & 0.694 & 0.546 & 0.510 \\

\rowcolor{lightgray}
\multicolumn{14}{l}{\emph{\textbf{GPT-J-6B PMI}}}\\
Random & 0.903 & 0.848 & 0.900 & 0.765 & 0.450 & 0.747 & 0.589 & 0.601 & 0.912 & 0.461 & 0.407 & 0.521 & 0.513 \\
LocalE & 0.903 & 0.848 & 0.898 & 0.752 & 0.445 & 0.788 & 0.602 & 0.608 & 0.908 & 0.468 & 0.581 & 0.535 & 0.580 \\
GlobalE & 0.895 & 0.831 & 0.879 & 0.773 & 0.436 & 0.799 & 0.629 & 0.590 & 0.891 & 0.459 & 0.581 & 0.532 & 0.551 \\
\hline
PDO (\settingA) & 0.919 & 0.848 & 0.900 & 0.780 & 0.450 & 0.773 & 0.574 & 0.606 & 0.908 & 0.471 & 0.416 & 0.538 & 0.507 \\
PDO (\settingB) & 0.903 & 0.848 & 0.899 & 0.753 & 0.445 & 0.792 & 0.644 & 0.608 & 0.910 & 0.471 & 0.591 & 0.540 & 0.568 \\
PDO (\settingC) & 0.926 & 0.883 & 0.906 & 0.805 & 0.466 & 0.830 & 0.644 & 0.616 & 0.922 & 0.500 & 0.614 & 0.547 & 0.593 \\
\hline
Oracle & 0.957 & 0.883 & 0.912 & 0.826 & 0.488 & 0.834 & 0.655 & 0.625 & 0.931 & 0.522 & 0.620 & 0.561 & 0.604 \\
\bottomrule
\end{tabular}
}
\label{tab:complete_classification_p2}
\vspace{0pt}
\end{table*}

\begin{table*}[t]
\centering
\caption{Complete classification results (measured by accuracy), Part 3.}
\vspace{0pt}
\resizebox{\textwidth}{!}{
\begin{tabular}{llrrrrrrrrrrrr}
\toprule
\, & SST2 & CR & MR & SUBJ & SST5 & AGNews & TREC & \begin{tabular}[c]{@{}l@{}l} Yahoo\\ Topics \end{tabular} & Dbpedia & FPB & \begin{tabular}[c]{@{}l@{}l} Tweet\\Offensive \end{tabular} & \begin{tabular}[c]{@{}l@{}l} Tweet\\ Irony \end{tabular} & \begin{tabular}[c]{@{}l@{}l} Tweet\\ Hate \end{tabular}
\\ 
\rowcolor{lightgray}
\multicolumn{14}{l}{\emph{\textbf{LLaMA-7B Direct}}}\\
Random & 0.928 & 0.901 & 0.913 & 0.594 & 0.471 & 0.849 & 0.607 & 0.527 & 0.813 & 0.648 & 0.683 & 0.538 & 0.558 \\
LocalE & 0.927 & 0.901 & 0.911 & 0.627 & 0.467 & 0.853 & 0.589 & 0.530 & 0.809 & 0.656 & 0.684 & 0.540 & 0.560 \\
GlobalE & 0.932 & 0.917 & 0.927 & 0.732 & 0.467 & 0.869 & 0.639 & 0.559 & 0.849 & 0.689 & 0.679 & 0.536 & 0.572 \\
\hline
PDO (\settingA) & 0.931 & 0.911 & 0.910 & 0.689 & 0.455 & 0.846 & 0.587 & 0.552 & 0.815 & 0.678 & 0.679 & 0.538 & 0.548 \\
PDO (\settingB) & 0.933 & 0.915 & 0.926 & 0.733 & 0.459 & 0.865 & 0.642 & 0.564 & 0.849 & 0.687 & 0.679 & 0.540 & 0.572 \\
PDO (\settingC) & 0.936 & 0.915 & 0.927 & 0.733 & 0.487 & 0.865 & 0.656 & 0.565 & 0.850 & 0.668 & 0.684 & 0.541 & 0.591 \\
\hline
Oracle & 0.940 & 0.922 & 0.930 & 0.735 & 0.507 & 0.874 & 0.668 & 0.569 & 0.854 & 0.690 & 0.684 & 0.542 & 0.598 \\

\rowcolor{lightgray}
\multicolumn{14}{l}{\emph{\textbf{LLaMA-7B PMI}}}\\
Random & 0.920 & 0.927 & 0.889 & 0.728 & 0.426 & 0.855 & 0.591 & 0.664 & 0.902 & 0.779 & 0.487 & 0.488 & 0.498 \\
LocalE & 0.922 & 0.918 & 0.904 & 0.652 & 0.434 & 0.855 & 0.637 & 0.658 & 0.896 & 0.782 & 0.567 & 0.491 & 0.485 \\
GlobalE & 0.936 & 0.932 & 0.914 & 0.808 & 0.453 & 0.878 & 0.649 & 0.670 & 0.930 & 0.762 & 0.564 & 0.518 & 0.548 \\
\hline
PDO (\settingA) & 0.925 & 0.921 & 0.917 & 0.720 & 0.424 & 0.855 & 0.614 & 0.670 & 0.916 & 0.787 & 0.599 & 0.488 & 0.518 \\
PDO (\settingB) & 0.931 & 0.917 & 0.912 & 0.800 & 0.454 & 0.870 & 0.659 & 0.666 & 0.925 & 0.795 & 0.561 & 0.516 & 0.552 \\
PDO (\settingC) & 0.931 & 0.917 & 0.912 & 0.803 & 0.447 & 0.866 & 0.681 & 0.670 & 0.929 & 0.787 & 0.652 & 0.518 & 0.579 \\
\hline
Oracle & 0.940 & 0.939 & 0.916 & 0.823 & 0.470 & 0.881 & 0.699 & 0.684 & 0.934 & 0.823 & 0.656 & 0.534 & 0.591 \\

\rowcolor{lightgray}
\multicolumn{14}{l}{\emph{\textbf{OPT-13B Direct}}}\\
Random & 0.871 & 0.905 & 0.868 & 0.672 & 0.485 & 0.835 & 0.393 & 0.541 & 0.837 & 0.684 & 0.494 & 0.476 & 0.558 \\
LocalE & 0.884 & 0.907 & 0.861 & 0.628 & 0.481 & 0.857 & 0.398 & 0.540 & 0.838 & 0.680 & 0.454 & 0.466 & 0.484 \\
GlobalE & 0.893 & 0.891 & 0.847 & 0.695 & 0.457 & 0.840 & 0.392 & 0.544 & 0.814 & 0.674 & 0.527 & 0.498 & 0.524 \\
\hline
PDO (\settingA) & 0.925 & 0.917 & 0.879 & 0.694 & 0.472 & 0.836 & 0.400 & 0.560 & 0.842 & 0.698 & 0.487 & 0.480 & 0.548 \\
PDO (\settingB) & 0.897 & 0.904 & 0.860 & 0.707 & 0.479 & 0.849 & 0.403 & 0.553 & 0.828 & 0.692 & 0.501 & 0.498 & 0.515 \\
PDO (\settingC) & 0.931 & 0.922 & 0.899 & 0.792 & 0.493 & 0.866 & 0.435 & 0.575 & 0.857 & 0.671 & 0.586 & 0.503 & 0.525 \\
\hline
Oracle & 0.935 & 0.925 & 0.901 & 0.813 & 0.512 & 0.872 & 0.457 & 0.580 & 0.861 & 0.715 & 0.590 & 0.506 & 0.534 \\

\rowcolor{lightgray}
\multicolumn{14}{l}{\emph{\textbf{OPT-13B PMI}}}\\
Random & 0.944 & 0.916 & 0.918 & 0.647 & 0.430 & 0.835 & 0.399 & 0.618 & 0.893 & 0.579 & 0.534 & 0.488 & 0.498 \\
LocalE & 0.941 & 0.915 & 0.911 & 0.650 & 0.443 & 0.842 & 0.422 & 0.623 & 0.895 & 0.630 & 0.552 & 0.475 & 0.513 \\
GlobalE & 0.913 & 0.895 & 0.889 & 0.656 & 0.441 & 0.837 & 0.402 & 0.610 & 0.869 & 0.627 & 0.492 & 0.534 & 0.583 \\
\hline
PDO (\settingA) & 0.949 & 0.911 & 0.913 & 0.692 & 0.418 & 0.838 & 0.410 & 0.623 & 0.895 & 0.640 & 0.533 & 0.488 & 0.518 \\
PDO (\settingB) & 0.942 & 0.916 & 0.911 & 0.688 & 0.459 & 0.843 & 0.432 & 0.623 & 0.895 & 0.649 & 0.578 & 0.505 & 0.558 \\
PDO (\settingC) & 0.944 & 0.922 & 0.911 & 0.747 & 0.462 & 0.857 & 0.448 & 0.628 & 0.902 & 0.656 & 0.652 & 0.536 & 0.582 \\
\hline
Oracle & 0.957 & 0.931 & 0.932 & 0.790 & 0.481 & 0.873 & 0.473 & 0.639 & 0.910 & 0.694 & 0.667 & 0.549 & 0.613 \\

\rowcolor{lightgray}
\multicolumn{14}{l}{\emph{\textbf{LLaMA-13B Direct}}}\\
Random & 0.936 & 0.894 & 0.927 & 0.724 & 0.495 & 0.848 & 0.628 & 0.519 & 0.885 & 0.675 & 0.646 & 0.510 & 0.511 \\
LocalE & 0.926 & 0.893 & 0.917 & 0.729 & 0.482 & 0.853 & 0.624 & 0.510 & 0.892 & 0.660 & 0.661 & 0.511 & 0.521 \\
GlobalE & 0.943 & 0.920 & 0.943 & 0.811 & 0.490 & 0.864 & 0.670 & 0.551 & 0.887 & 0.735 & 0.562 & 0.539 & 0.560 \\
\hline
PDO (\settingA) & 0.936 & 0.894 & 0.930 & 0.751 & 0.492 & 0.849 & 0.608 & 0.526 & 0.898 & 0.703 & 0.624 & 0.525 & 0.521 \\
PDO (\settingB) & 0.946 & 0.918 & 0.943 & 0.804 & 0.488 & 0.862 & 0.679 & 0.551 & 0.909 & 0.735 & 0.560 & 0.545 & 0.564 \\
PDO (\settingC) & 0.948 & 0.917 & 0.940 & 0.788 & 0.502 & 0.860 & 0.697 & 0.552 & 0.909 & 0.719 & 0.684 & 0.540 & 0.565 \\
\hline
Oracle & 0.951 & 0.921 & 0.945 & 0.844 & 0.512 & 0.867 & 0.706 & 0.559 & 0.911 & 0.768 & 0.685 & 0.555 & 0.594 \\

\rowcolor{lightgray}
\multicolumn{14}{l}{\emph{\textbf{LLaMA-13B PMI}}}\\
Random & 0.932 & 0.928 & 0.928 & 0.767 & 0.426 & 0.851 & 0.634 & 0.655 & 0.942 & 0.686 & 0.629 & 0.511 & 0.501 \\
LocalE & 0.932 & 0.913 & 0.926 & 0.681 & 0.435 & 0.863 & 0.622 & 0.653 & 0.905 & 0.713 & 0.638 & 0.494 & 0.514 \\
GlobalE & 0.947 & 0.938 & 0.940 & 0.840 & 0.459 & 0.872 & 0.655 & 0.670 & 0.921 & 0.697 & 0.552 & 0.541 & 0.545 \\
\hline
PDO (\settingA) & 0.934 & 0.928 & 0.930 & 0.755 & 0.432 & 0.847 & 0.622 & 0.647 & 0.943 & 0.739 & 0.618 & 0.504 & 0.464 \\
PDO (\settingB) & 0.943 & 0.921 & 0.933 & 0.825 & 0.449 & 0.868 & 0.650 & 0.660 & 0.946 & 0.740 & 0.569 & 0.542 & 0.558 \\
PDO (\settingC) & 0.942 & 0.921 & 0.934 & 0.823 & 0.460 & 0.861 & 0.701 & 0.665 & 0.946 & 0.766 & 0.681 & 0.546 & 0.574 \\
\hline
Oracle & 0.949 & 0.942 & 0.941 & 0.858 & 0.477 & 0.877 & 0.720 & 0.678 & 0.955 & 0.801 & 0.688 & 0.576 & 0.600 \\
\bottomrule
\end{tabular}
}
\label{tab:complete_classification_p3}
\vspace{0pt}
\end{table*}

\begin{table*}[t]
\centering
\caption{Complete confidence calibration results (measured by ECE), Part 1.}
\vspace{0pt}
\resizebox{\textwidth}{!}{
\begin{tabular}{llrrrrrrrrrrrr}
\toprule
\, & SST2 & CR & MR & SUBJ & SST5 & AGNews & TREC & \begin{tabular}[c]{@{}l@{}l} Yahoo\\ Topics \end{tabular} & Dbpedia & FPB & \begin{tabular}[c]{@{}l@{}l} Tweet\\Offensive \end{tabular} & \begin{tabular}[c]{@{}l@{}l} Tweet\\ Irony \end{tabular} & \begin{tabular}[c]{@{}l@{}l} Tweet\\ Hate \end{tabular}
\\ 
\rowcolor{lightgray}
\multicolumn{14}{l}{\emph{\textbf{GPT2-large Direct}}}\\
Random & 0.242 & 0.314 & 0.240 & 0.254 & 0.273 & 0.268 & 0.325 & 0.246 & 0.263 & 0.319 & 0.289 & 0.265 & 0.292 \\
LocalE & 0.250 & 0.221 & 0.232 & 0.193 & 0.209 & 0.223 & 0.272 & 0.208 & 0.225 & 0.257 & 0.164 & 0.129 & 0.175 \\
GlobalE & 0.278 & 0.224 & 0.235 & 0.197 & 0.180 & 0.216 & 0.27 & 0.216 & 0.19 & 0.256 & 0.179 & 0.136 & 0.204 \\ 
\hline
PDO (\settingA) & 0.260 & 0.226 & 0.208 & 0.264 & 0.224 & 0.237 & 0.301 & 0.215 & 0.213 & 0.242 & 0.241 & 0.163 & 0.185 \\
PDO (\settingB) & 0.247 & 0.221 & 0.23 & 0.195 & 0.21 & 0.223 & 0.275 & 0.208 & 0.225 & 0.258 & 0.160 & 0.128 & 0.175 \\
PDO (\settingC) & 0.275 & 0.223 & 0.233 & 0.197 & 0.202 & 0.220 & 0.275 & 0.231 & 0.195 & 0.243 & 0.125 & 0.129 & 0.183 \\ 
\hline
Oracle & 0.028 & 0.127 & 0.023 & 0.107 & 0.054 & 0.140 & 0.097 & 0.085 & 0.166 & 0.157 & 0.08 & 0.027 & 0.082 \\

\rowcolor{lightgray}
\multicolumn{14}{l}{\emph{\textbf{GPT2-large PMI}}}\\
Random & 0.243 & 0.224 & 0.221 & 0.205 & 0.150 & 0.225 & 0.340 & 0.262 & 0.222 & 0.238 & 0.245 & 0.141 & 0.132 \\
LocalE & 0.250 & 0.221 & 0.232 & 0.193 & 0.209 & 0.223 & 0.272 & 0.208 & 0.225 & 0.257 & 0.164 & 0.129 & 0.175 \\
GlobalE & 0.278 & 0.224 & 0.235 & 0.197 & 0.180 & 0.216 & 0.270 & 0.216 & 0.19 & 0.256 & 0.179 & 0.136 & 0.204 \\ 
\hline
PDO (\settingA) & 0.264 & 0.217 & 0.218 & 0.240 & 0.144 & 0.209 & 0.278 & 0.235 & 0.217 & 0.254 & 0.226 & 0.142 & 0.122 \\
PDO (\settingB) & 0.247 & 0.221 & 0.230 & 0.195 & 0.210 & 0.223 & 0.275 & 0.208 & 0.225 & 0.258 & 0.160 & 0.128 & 0.175 \\
PDO (\settingC) & 0.261 & 0.228 & 0.232 & 0.197 & 0.157 & 0.221 & 0.233 & 0.21 & 0.221 & 0.262 & 0.136 & 0.108 & 0.109 \\ 
\hline
Oracle & 0.272 & 0.224 & 0.236 & 0.213 & 0.187 & 0.234 & 0.254 & 0.235 & 0.211 & 0.314 & 0.138 & 0.124 & 0.105 \\

\rowcolor{lightgray}
\multicolumn{14}{l}{\emph{\textbf{GPT2-xl Direct}}}\\
Random & 0.277 & 0.289 & 0.264 & 0.192 & 0.192 & 0.179 & 0.27 & 0.230 & 0.168 & 0.258 & 0.348 & 0.199 & 0.265 \\
LocalE & 0.215 & 0.23 & 0.205 & 0.173 & 0.151 & 0.179 & 0.225 & 0.213 & 0.177 & 0.216 & 0.162 & 0.121 & 0.158 \\
GlobalE & 0.237 & 0.198 & 0.196 & 0.172 & 0.138 & 0.168 & 0.236 & 0.213 & 0.16 & 0.217 & 0.162 & 0.125 & 0.160 \\
\hline
PDO (\settingA) & 0.240 & 0.216 & 0.209 & 0.158 & 0.177 & 0.169 & 0.239 & 0.216 & 0.165 & 0.248 & 0.289 & 0.190 & 0.207 \\
PDO (\settingB) & 0.213 & 0.233 & 0.205 & 0.170 & 0.148 & 0.179 & 0.242 & 0.213 & 0.177 & 0.217 & 0.163 & 0.122 & 0.157 \\
PDO (\settingC) & 0.239 & 0.212 & 0.200 & 0.165 & 0.157 & 0.164 & 0.242 & 0.212 & 0.159 & 0.239 & 0.165 & 0.121 & 0.151 \\
\hline
Oracle & 0.237 & 0.199 & 0.197 & 0.182 & 0.136 & 0.174 & 0.247 & 0.213 & 0.16 & 0.233 & 0.173 & 0.196 & 0.187 \\

\rowcolor{lightgray}
\multicolumn{14}{l}{\emph{\textbf{GPT2-xl PMI}}}\\
Random & 0.248 & 0.218 & 0.214 & 0.191 & 0.228 & 0.184 & 0.304 & 0.345 & 0.197 & 0.263 & 0.297 & 0.234 & 0.172 \\
LocalE & 0.258 & 0.233 & 0.226 & 0.172 & 0.149 & 0.177 & 0.237 & 0.322 & 0.198 & 0.231 & 0.202 & 0.153 & 0.123 \\
GlobalE & 0.265 & 0.224 & 0.245 & 0.181 & 0.162 & 0.199 & 0.252 & 0.330 & 0.211 & 0.234 & 0.21 & 0.155 & 0.149 \\
\hline
PDO (\settingA) & 0.252 & 0.221 & 0.187 & 0.159 & 0.174 & 0.165 & 0.271 & 0.333 & 0.173 & 0.255 & 0.28 & 0.186 & 0.196 \\
PDO (\settingB) & 0.255 & 0.234 & 0.224 & 0.174 & 0.147 & 0.177 & 0.246 & 0.322 & 0.198 & 0.233 & 0.204 & 0.159 & 0.125 \\
PDO (\settingC) & 0.284 & 0.239 & 0.241 & 0.176 & 0.153 & 0.173 & 0.246 & 0.325 & 0.166 & 0.236 & 0.204 & 0.153 & 0.115 \\
\hline
Oracle & 0.287 & 0.248 & 0.241 & 0.182 & 0.173 & 0.185 & 0.244 & 0.333 & 0.175 & 0.235 & 0.200 & 0.183 & 0.128 \\

\rowcolor{lightgray}
\multicolumn{14}{l}{\emph{\textbf{OPT-1.3B Direct}}}\\
Random & 0.163 & 0.126 & 0.147 & 0.217 & 0.240 & 0.167 & 0.318 & 0.229 & 0.133 & 0.220 & 0.270 & 0.349 & 0.241 \\
LocalE & 0.165 & 0.156 & 0.159 & 0.189 & 0.281 & 0.173 & 0.256 & 0.194 & 0.139 & 0.214 & 0.157 & 0.181 & 0.165 \\
GlobalE & 0.177 & 0.155 & 0.148 & 0.194 & 0.272 & 0.187 & 0.271 & 0.204 & 0.148 & 0.226 & 0.155 & 0.192 & 0.181 \\
\hline
PDO (\settingA) & 0.157 & 0.135 & 0.152 & 0.209 & 0.213 & 0.171 & 0.319 & 0.216 & 0.130 & 0.210 & 0.193 & 0.225 & 0.169 \\
PDO (\settingB) & 0.165 & 0.148 & 0.148 & 0.189 & 0.284 & 0.169 & 0.269 & 0.194 & 0.139 & 0.213 & 0.143 & 0.175 & 0.167 \\
PDO (\settingC) & 0.169 & 0.133 & 0.147 & 0.192 & 0.205 & 0.169 & 0.269 & 0.21 & 0.129 & 0.195 & 0.121 & 0.181 & 0.152 \\
\hline
Oracle & 0.169 & 0.132 & 0.139 & 0.200 & 0.212 & 0.162 & 0.270 & 0.205 & 0.127 & 0.227 & 0.157 & 0.270 & 0.244 \\

\rowcolor{lightgray}
\multicolumn{14}{l}{\emph{\textbf{OPT-1.3B PMI}}}\\
Random & 0.180 & 0.171 & 0.173 & 0.230 & 0.284 & 0.196 & 0.460 & 0.294 & 0.143 & 0.206 & 0.192 & 0.168 & 0.167 \\
LocalE & 0.187 & 0.181 & 0.183 & 0.155 & 0.215 & 0.187 & 0.342 & 0.244 & 0.151 & 0.237 & 0.144 & 0.112 & 0.127 \\
GlobalE & 0.182 & 0.192 & 0.185 & 0.149 & 0.230 & 0.195 & 0.355 & 0.241 & 0.159 & 0.244 & 0.145 & 0.112 & 0.135 \\ 
\hline
PDO (\settingA) & 0.181 & 0.175 & 0.177 & 0.194 & 0.300 & 0.173 & 0.370 & 0.269 & 0.131 & 0.231 & 0.190 & 0.161 & 0.165 \\
PDO (\settingB) & 0.179 & 0.185 & 0.187 & 0.152 & 0.218 & 0.189 & 0.345 & 0.244 & 0.151 & 0.238 & 0.140 & 0.103 & 0.127 \\
PDO (\settingC) & 0.195 & 0.171 & 0.180 & 0.154 & 0.225 & 0.180 & 0.345 & 0.265 & 0.127 & 0.204 & 0.126 & 0.112 & 0.128 \\
\hline
Oracle & 0.193 & 0.178 & 0.172 & 0.159 & 0.213 & 0.190 & 0.386 & 0.297 & 0.127 & 0.262 & 0.139 & 0.169 & 0.139 \\
\bottomrule
\end{tabular}
}
\label{tab:complete_calibration_p1}
\vspace{0pt}
\end{table*}

\begin{table*}[t]
\centering
\caption{
Complete confidence calibration results (measured by ECE), Part 2.}
\vspace{0pt}
\resizebox{\textwidth}{!}{
\begin{tabular}{llrrrrrrrrrrrr}
\toprule
\, & SST2 & CR & MR & SUBJ & SST5 & AGNews & TREC & \begin{tabular}[c]{@{}l@{}l} Yahoo\\ Topics \end{tabular} & Dbpedia & FPB & \begin{tabular}[c]{@{}l@{}l} Tweet\\Offensive \end{tabular} & \begin{tabular}[c]{@{}l@{}l} Tweet\\ Irony \end{tabular} & \begin{tabular}[c]{@{}l@{}l} Tweet\\ Hate \end{tabular}
\\ 
\rowcolor{lightgray}
\multicolumn{14}{l}{\emph{\textbf{OPT-2.7B Direct}}}\\
Random & 0.204 & 0.131 & 0.153 & 0.226 & 0.218 & 0.177 & 0.276 & 0.230 & 0.113 & 0.269 & 0.260 & 0.313 & 0.251 \\
LocalE & 0.234 & 0.148 & 0.174 & 0.195 & 0.197 & 0.191 & 0.232 & 0.211 & 0.126 & 0.282 & 0.174 & 0.163 & 0.140 \\
GlobalE & 0.216 & 0.145 & 0.165 & 0.207 & 0.188 & 0.189 & 0.226 & 0.218 & 0.106 & 0.280 & 0.166 & 0.178 & 0.137 \\
\hline
PDO (\settingA) & 0.211 & 0.143 & 0.158 & 0.199 & 0.203 & 0.182 & 0.278 & 0.228 & 0.116 & 0.282 & 0.192 & 0.189 & 0.158 \\
PDO (\settingB) & 0.235 & 0.148 & 0.173 & 0.199 & 0.196 & 0.193 & 0.237 & 0.211 & 0.126 & 0.282 & 0.179 & 0.160 & 0.142 \\
PDO (\settingC) & 0.223 & 0.145 & 0.165 & 0.206 & 0.199 & 0.184 & 0.237 & 0.224 & 0.105 & 0.260 & 0.143 & 0.163 & 0.141 \\
\hline
Oracle & 0.218 & 0.146 & 0.160 & 0.209 & 0.213 & 0.185 & 0.235 & 0.222 & 0.107 & 0.305 & 0.204 & 0.202 & 0.144 \\

\rowcolor{lightgray}
\multicolumn{14}{l}{\emph{\textbf{OPT-2.7B PMI}}}\\
Random & 0.229 & 0.176 & 0.185 & 0.200 & 0.247 & 0.208 & 0.510 & 0.322 & 0.156 & 0.248 & 0.182 & 0.159 & 0.147 \\
LocalE & 0.250 & 0.190 & 0.201 & 0.200 & 0.184 & 0.187 & 0.403 & 0.281 & 0.177 & 0.250 & 0.137 & 0.106 & 0.112 \\
GlobalE & 0.255 & 0.191 & 0.199 & 0.205 & 0.189 & 0.201 & 0.405 & 0.302 & 0.195 & 0.255 & 0.145 & 0.131 & 0.135 \\
\hline
PDO (\settingA) & 0.227 & 0.184 & 0.187 & 0.198 & 0.204 & 0.177 & 0.485 & 0.299 & 0.127 & 0.267 & 0.159 & 0.143 & 0.144 \\
PDO (\settingB) & 0.251 & 0.186 & 0.202 & 0.198 & 0.182 & 0.192 & 0.393 & 0.281 & 0.177 & 0.243 & 0.134 & 0.112 & 0.113 \\
PDO (\settingC) & 0.242 & 0.186 & 0.191 & 0.214 & 0.187 & 0.181 & 0.393 & 0.300 & 0.119 & 0.258 & 0.133 & 0.106 & 0.123 \\
\hline
Oracle & 0.234 & 0.181 & 0.194 & 0.217 & 0.218 & 0.184 & 0.425 & 0.310 & 0.125 & 0.274 & 0.142 & 0.117 & 0.120 \\

\rowcolor{lightgray}
\multicolumn{14}{l}{\emph{\textbf{OPT-6.7B Direct}}}\\
Random & 0.142 & 0.120 & 0.111 & 0.199 & 0.253 & 0.174 & 0.269 & 0.201 & 0.103 & 0.237 & 0.195 & 0.207 & 0.238 \\
LocalE & 0.162 & 0.128 & 0.124 & 0.185 & 0.232 & 0.171 & 0.246 & 0.190 & 0.108 & 0.280 & 0.127 & 0.107 & 0.099 \\
GlobalE & 0.155 & 0.126 & 0.120 & 0.192 & 0.228 & 0.150 & 0.262 & 0.197 & 0.097 & 0.261 & 0.137 & 0.117 & 0.098 \\
\hline
PDO (\settingA) & 0.147 & 0.120 & 0.111 & 0.189 & 0.237 & 0.171 & 0.273 & 0.195 & 0.103 & 0.243 & 0.193 & 0.173 & 0.193 \\
PDO (\settingB) & 0.161 & 0.129 & 0.124 & 0.188 & 0.233 & 0.170 & 0.250 & 0.190 & 0.108 & 0.280 & 0.128 & 0.103 & 0.095 \\
PDO (\settingC) & 0.144 & 0.169 & 0.112 & 0.192 & 0.228 & 0.150 & 0.250 & 0.197 & 0.095 & 0.217 & 0.131 & 0.107 & 0.117 \\
\hline
Oracle & 0.149 & 0.125 & 0.111 & 0.200 & 0.256 & 0.150 & 0.255 & 0.193 & 0.099 & 0.262 & 0.162 & 0.178 & 0.195 \\

\rowcolor{lightgray}
\multicolumn{14}{l}{\emph{\textbf{OPT-6.7B PMI}}}\\
Random & 0.187 & 0.166 & 0.157 & 0.212 & 0.274 & 0.209 & 0.406 & 0.282 & 0.156 & 0.262 & 0.263 & 0.148 & 0.188 \\
LocalE & 0.194 & 0.171 & 0.166 & 0.189 & 0.220 & 0.190 & 0.299 & 0.262 & 0.172 & 0.299 & 0.123 & 0.093 & 0.107 \\
GlobalE & 0.205 & 0.177 & 0.192 & 0.195 & 0.205 & 0.188 & 0.313 & 0.271 & 0.175 & 0.305 & 0.134 & 0.101 & 0.111 \\
\hline
PDO (\settingA) & 0.186 & 0.168 & 0.157 & 0.185 & 0.243 & 0.204 & 0.321 & 0.278 & 0.130 & 0.289 & 0.232 & 0.143 & 0.220 \\
PDO (\settingB) & 0.196 & 0.170 & 0.166 & 0.186 & 0.219 & 0.190 & 0.305 & 0.260 & 0.172 & 0.299 & 0.129 & 0.094 & 0.108 \\
PDO (\settingC) & 0.191 & 0.125 & 0.155 & 0.183 & 0.239 & 0.187 & 0.305 & 0.265 & 0.128 & 0.304 & 0.124 & 0.093 & 0.113 \\
\hline
Oracle & 0.190 & 0.167 & 0.160 & 0.186 & 0.237 & 0.190 & 0.313 & 0.273 & 0.132 & 0.282 & 0.126 & 0.094 & 0.125 \\

\rowcolor{lightgray}
\multicolumn{14}{l}{\emph{\textbf{GPT-J-6B Direct}}}\\
Random & 0.221 & 0.154 & 0.144 & 0.183 & 0.246 & 0.142 & 0.230 & 0.212 & 0.109 & 0.177 & 0.204 & 0.252 & 0.291 \\
LocalE & 0.243 & 0.166 & 0.162 & 0.183 & 0.221 & 0.148 & 0.203 & 0.196 & 0.118 & 0.157 & 0.124 & 0.142 & 0.209 \\
GlobalE & 0.247 & 0.167 & 0.148 & 0.178 & 0.224 & 0.142 & 0.198 & 0.200 & 0.102 & 0.144 & 0.124 & 0.171 & 0.254 \\
\hline
PDO (\settingA) & 0.243 & 0.149 & 0.155 & 0.162 & 0.231 & 0.145 & 0.231 & 0.208 & 0.104 & 0.166 & 0.183 & 0.144 & 0.225 \\
PDO (\settingB) & 0.242 & 0.166 & 0.160 & 0.184 & 0.221 & 0.148 & 0.205 & 0.192 & 0.117 & 0.154 & 0.120 & 0.145 & 0.202 \\
PDO (\settingC) & 0.252 & 0.199 & 0.153 & 0.180 & 0.225 & 0.136 & 0.205 & 0.203 & 0.101 & 0.175 & 0.127 & 0.142 & 0.212 \\
\hline
Oracle & 0.254 & 0.167 & 0.151 & 0.186 & 0.228 & 0.137 & 0.195 & 0.204 & 0.101 & 0.189 & 0.147 & 0.180 & 0.260 \\

\rowcolor{lightgray}
\multicolumn{14}{l}{\emph{\textbf{GPT-J-6B PMI}}}\\
Random & 0.247 & 0.176 & 0.182 & 0.172 & 0.218 & 0.256 & 0.252 & 0.317 & 0.126 & 0.176 & 0.227 & 0.088 & 0.139 \\
LocalE & 0.265 & 0.189 & 0.193 & 0.179 & 0.202 & 0.218 & 0.206 & 0.290 & 0.140 & 0.154 & 0.101 & 0.066 & 0.097 \\
GlobalE & 0.275 & 0.199 & 0.187 & 0.175 & 0.213 & 0.234 & 0.201 & 0.292 & 0.145 & 0.149 & 0.095 & 0.075 & 0.083 \\
\hline
PDO (\settingA) & 0.255 & 0.177 & 0.182 & 0.177 & 0.202 & 0.229 & 0.229 & 0.298 & 0.115 & 0.179 & 0.213 & 0.085 & 0.126 \\
PDO (\settingB) & 0.263 & 0.190 & 0.190 & 0.180 & 0.202 & 0.213 & 0.210 & 0.288 & 0.140 & 0.154 & 0.100 & 0.071 & 0.088 \\
PDO (\settingC) & 0.263 & 0.168 & 0.187 & 0.190 & 0.203 & 0.200 & 0.210 & 0.293 & 0.109 & 0.164 & 0.106 & 0.066 & 0.113 \\
\hline
Oracle & 0.261 & 0.196 & 0.186 & 0.195 & 0.214 & 0.200 & 0.214 & 0.303 & 0.114 & 0.175 & 0.107 & 0.084 & 0.114 \\
\bottomrule
\end{tabular}
}
\label{tab:complete_calibration_p2}
\vspace{0pt}
\end{table*}

\begin{table*}[t]
\centering
\caption{Complete confidence calibration results (measured by ECE), Part 3.}
\vspace{0pt}
\resizebox{\textwidth}{!}{
\begin{tabular}{llrrrrrrrrrrrr}
\toprule
\, & SST2 & CR & MR & SUBJ & SST5 & AGNews & TREC & \begin{tabular}[c]{@{}l@{}l} Yahoo\\ Topics \end{tabular} & Dbpedia & FPB & \begin{tabular}[c]{@{}l@{}l} Tweet\\Offensive \end{tabular} & \begin{tabular}[c]{@{}l@{}l} Tweet\\ Irony \end{tabular} & \begin{tabular}[c]{@{}l@{}l} Tweet\\ Hate \end{tabular}
\\ 
\rowcolor{lightgray}
\multicolumn{14}{l}{\emph{\textbf{LLaMA-7B Direct}}}\\
Random & 0.122 & 0.094 & 0.091 & 0.269 & 0.263 & 0.122 & 0.205 & 0.222 & 0.140 & 0.224 & 0.130 & 0.217 & 0.175 \\
LocalE & 0.142 & 0.102 & 0.100 & 0.218 & 0.253 & 0.127 & 0.192 & 0.209 & 0.147 & 0.206 & 0.107 & 0.192 & 0.153 \\
GlobalE & 0.118 & 0.088 & 0.085 & 0.155 & 0.245 & 0.119 & 0.187 & 0.197 & 0.130 & 0.202 & 0.138 & 0.177 & 0.147 \\
\hline
PDO (\settingA) & 0.125 & 0.09 & 0.094 & 0.164 & 0.261 & 0.119 & 0.216 & 0.204 & 0.139 & 0.199 & 0.108 & 0.158 & 0.130 \\
PDO (\settingB) & 0.118 & 0.09 & 0.087 & 0.157 & 0.249 & 0.118 & 0.185 & 0.199 & 0.128 & 0.197 & 0.109 & 0.141 & 0.117 \\
PDO (\settingC) & 0.121 & 0.09 & 0.088 & 0.157 & 0.239 & 0.117 & 0.186 & 0.203 & 0.127 & 0.189 & 0.099 & 0.140 & 0.126 \\
\hline
Oracle & 0.125 & 0.085 & 0.087 & 0.155 & 0.255 & 0.117 & 0.181 & 0.209 & 0.131 & 0.201 & 0.134 & 0.200 & 0.172 \\

\rowcolor{lightgray}
\multicolumn{14}{l}{\emph{\textbf{LLaMA-7B PMI}}}\\
Random & 0.160 & 0.151 & 0.141 & 0.170 & 0.243 & 0.161 & 0.318 & 0.241 & 0.135 & 0.277 & 0.151 & 0.123 & 0.134 \\
LocalE & 0.171 & 0.152 & 0.145 & 0.190 & 0.209 & 0.161 & 0.229 & 0.234 & 0.160 & 0.292 & 0.107 & 0.096 & 0.111 \\
GlobalE & 0.165 & 0.149 & 0.133 & 0.170 & 0.245 & 0.170 & 0.250 & 0.236 & 0.105 & 0.265 & 0.082 & 0.084 & 0.084 \\
\hline
PDO (\settingA) & 0.164 & 0.149 & 0.137 & 0.157 & 0.228 & 0.142 & 0.267 & 0.239 & 0.198 & 0.276 & 0.105 & 0.115 & 0.121 \\
PDO (\settingB) & 0.164 & 0.149 & 0.140 & 0.171 & 0.232 & 0.154 & 0.236 & 0.233 & 0.109 & 0.289 & 0.092 & 0.080 & 0.082 \\
PDO (\settingC) & 0.164 & 0.149 & 0.140 & 0.171 & 0.220 & 0.142 & 0.220 & 0.236 & 0.102 & 0.283 & 0.123 & 0.084 & 0.094 \\
\hline
Oracle & 0.167 & 0.151 & 0.138 & 0.173 & 0.223 & 0.167 & 0.224 & 0.244 & 0.105 & 0.307 & 0.130 & 0.087 & 0.098 \\

\rowcolor{lightgray}
\multicolumn{14}{l}{\emph{\textbf{OPT-13B Direct}}}\\
Random & 0.132 & 0.100 & 0.110 & 0.196 & 0.248 & 0.130 & 0.240 & 0.211 & 0.115 & 0.224 & 0.282 & 0.379 & 0.280 \\
LocalE & 0.140 & 0.105 & 0.116 & 0.155 & 0.237 & 0.129 & 0.215 & 0.199 & 0.112 & 0.237 & 0.290 & 0.401 & 0.248 \\
GlobalE & 0.137 & 0.097 & 0.101 & 0.201 & 0.238 & 0.128 & 0.218 & 0.189 & 0.105 & 0.217 & 0.173 & 0.235 & 0.204 \\
\hline
PDO (\settingA) & 0.134 & 0.100 & 0.109 & 0.177 & 0.243 & 0.128 & 0.234 & 0.197 & 0.111 & 0.220 & 0.215 & 0.253 & 0.212 \\
PDO (\settingB) & 0.141 & 0.106 & 0.116 & 0.159 & 0.236 & 0.131 & 0.207 & 0.198 & 0.121 & 0.217 & 0.191 & 0.219 & 0.184 \\
PDO (\settingC) & 0.135 & 0.094 & 0.104 & 0.198 & 0.227 & 0.127 & 0.214 & 0.187 & 0.104 & 0.200 & 0.170 & 0.219 & 0.180 \\
\hline
Oracle & 0.138 & 0.098 & 0.101 & 0.211 & 0.242 & 0.125 & 0.200 & 0.197 & 0.102 & 0.219 & 0.186 & 0.248 & 0.201 \\

\rowcolor{lightgray}
\multicolumn{14}{l}{\emph{\textbf{OPT-13B PMI}}}\\
Random & 0.189 & 0.154 & 0.163 & 0.186 & 0.258 & 0.153 & 0.361 & 0.304 & 0.130 & 0.208 & 0.159 & 0.180 & 0.169 \\
LocalE & 0.202 & 0.156 & 0.166 & 0.133 & 0.223 & 0.145 & 0.254 & 0.281 & 0.140 & 0.225 & 0.133 & 0.149 & 0.134 \\
GlobalE & 0.185 & 0.155 & 0.163 & 0.190 & 0.235 & 0.157 & 0.276 & 0.290 & 0.121 & 0.201 & 0.114 & 0.114 & 0.106 \\
\hline
PDO (\settingA) & 0.190 & 0.152 & 0.160 & 0.179 & 0.249 & 0.141 & 0.299 & 0.288 & 0.111 & 0.216 & 0.156 & 0.180 & 0.167 \\
PDO (\settingB) & 0.203 & 0.156 & 0.163 & 0.146 & 0.209 & 0.147 & 0.243 & 0.279 & 0.139 & 0.223 & 0.115 & 0.110 & 0.105 \\
PDO (\settingC) & 0.196 & 0.155 & 0.161 & 0.194 & 0.221 & 0.140 & 0.252 & 0.282 & 0.112 & 0.216 & 0.130 & 0.110 & 0.117 \\
\hline
Oracle & 0.192 & 0.154 & 0.163 & 0.206 & 0.24 & 0.156 & 0.295 & 0.282 & 0.118 & 0.229 & 0.134 & 0.117 & 0.138 \\

\rowcolor{lightgray}
\multicolumn{14}{l}{\emph{\textbf{LLaMA-13B Direct}}}\\
Random & 0.101 & 0.115 & 0.093 & 0.205 & 0.251 & 0.121 & 0.184 & 0.213 & 0.112 & 0.220 & 0.137 & 0.153 & 0.186 \\
LocalE & 0.113 & 0.128 & 0.104 & 0.185 & 0.247 & 0.119 & 0.184 & 0.207 & 0.121 & 0.247 & 0.106 & 0.110 & 0.137 \\
GlobalE & 0.098 & 0.121 & 0.088 & 0.192 & 0.239 & 0.116 & 0.175 & 0.194 & 0.115 & 0.233 & 0.124 & 0.093 & 0.089 \\
\hline
PDO (\settingA) & 0.100 & 0.124 & 0.091 & 0.184 & 0.242 & 0.123 & 0.191 & 0.216 & 0.107 & 0.221 & 0.117 & 0.099 & 0.104 \\
PDO (\settingB) & 0.101 & 0.121 & 0.087 & 0.191 & 0.235 & 0.115 & 0.171 & 0.198 & 0.101 & 0.241 & 0.110 & 0.081 & 0.093 \\
PDO (\settingC) & 0.093 & 0.122 & 0.092 & 0.181 & 0.236 & 0.116 & 0.176 & 0.197 & 0.101 & 0.213 & 0.095 & 0.085 & 0.096 \\
\hline
Oracle & 0.096 & 0.120 & 0.089 & 0.203 & 0.244 & 0.115 & 0.179 & 0.197 & 0.104 & 0.255 & 0.137 & 0.139 & 0.127 \\
\rowcolor{lightgray}
\multicolumn{14}{l}{\emph{\textbf{LLaMA-13B PMI}}}\\
Random & 0.154 & 0.170 & 0.152 & 0.187 & 0.283 & 0.178 & 0.277 & 0.262 & 0.096 & 0.239 & 0.129 & 0.087 & 0.106 \\
LocalE & 0.161 & 0.178 & 0.161 & 0.197 & 0.235 & 0.156 & 0.256 & 0.238 & 0.098 & 0.253 & 0.137 & 0.075 & 0.106 \\
GlobalE & 0.158 & 0.175 & 0.156 & 0.204 & 0.258 & 0.174 & 0.239 & 0.242 & 0.103 & 0.230 & 0.105 & 0.069 & 0.082 \\
\hline
PDO (\settingA) & 0.152 & 0.172 & 0.151 & 0.191 & 0.264 & 0.151 & 0.261 & 0.257 & 0.079 & 0.257 & 0.131 & 0.081 & 0.109 \\
PDO (\settingB) & 0.159 & 0.176 & 0.150 & 0.206 & 0.247 & 0.153 & 0.241 & 0.234 & 0.081 & 0.276 & 0.103 & 0.074 & 0.078 \\
PDO (\settingC) & 0.159 & 0.176 & 0.157 & 0.200 & 0.246 & 0.146 & 0.220 & 0.239 & 0.079 & 0.253 & 0.115 & 0.079 & 0.088 \\
\hline
Oracle & 0.156 & 0.168 & 0.154 & 0.211 & 0.256 & 0.176 & 0.223 & 0.250 & 0.091 & 0.292 & 0.144 & 0.074 & 0.098 \\
\bottomrule
\end{tabular}
}
\label{tab:complete_calibration_p3}
\vspace{0pt}
\end{table*}

\end{document}